\RequirePackage{fix-cm}
\documentclass[smallextended]{svjour3}       
\smartqed  
\usepackage{graphicx}
\usepackage{booktabs} 
\usepackage{amsmath,amssymb,amsfonts}
\usepackage[algoruled, lined, linesnumbered]{algorithm2e}
\usepackage{algcompatible}
\usepackage{caption}
\newsavebox\CBox
\def\textBF#1{\sbox\CBox{#1}\resizebox{\wd\CBox}{\ht\CBox}{\textbf{#1}}}
\usepackage{textcomp}
\usepackage{hyperref}
\usepackage{longtable}
\usepackage{multirow}
\usepackage{placeins}
\usepackage{multicol}
\usepackage{subfig}
\usepackage{wasysym}
\usepackage{enumitem}
\usepackage{cases}
\usepackage{cancel}

\usepackage{cuted}
\usepackage{color,soul}
\usepackage{soulutf8}
\usepackage{comment}

\usepackage{etoolbox}
\newcommand*{\affaddr}[1]{#1}
\newcommand*{\affmark}[1][*]{\textsuperscript{#1}}

\newenvironment{MyColorPar}[1]{\leavevmode\color{#1}\ignorespaces}{}

\begin{document}

\title{Efficiently solving the thief orienteering problem with a max-min ant colony optimization approach}

\titlerunning{Efficiently solving the ThOP with a max-min ACO approach}        

\author{\mbox{Jonatas B. C. Chagas \protect\affmark[1,2,*]\href{https://orcid.org/0000-0001-7965-8419}{\includegraphics[width=10pt,height=10pt]{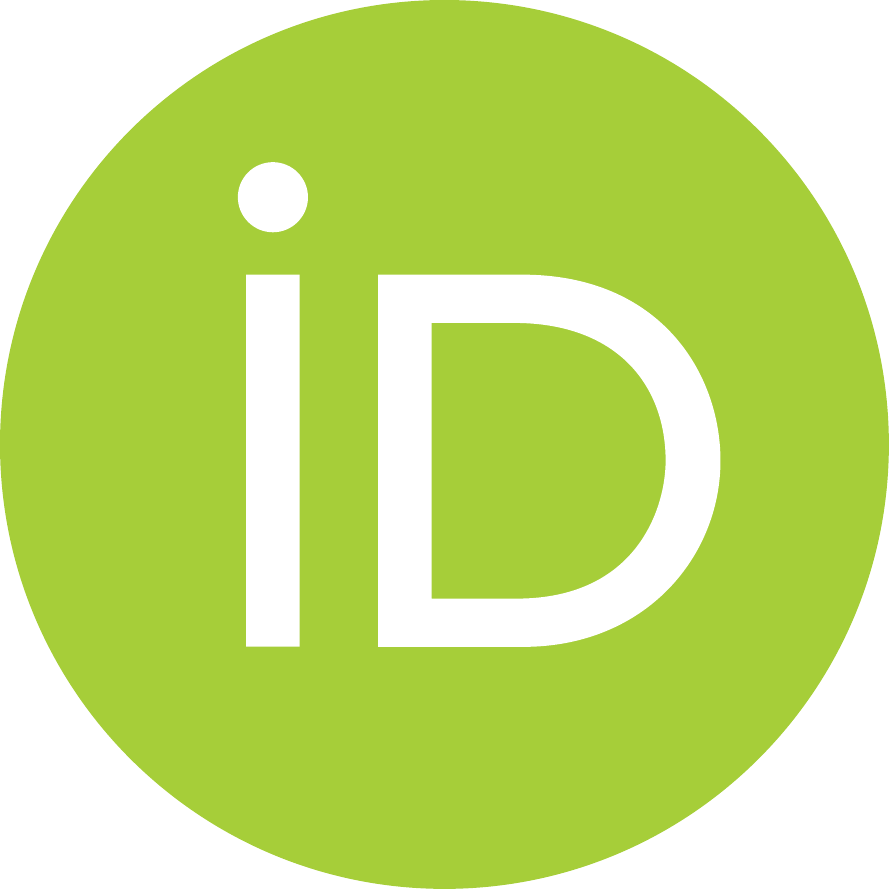}} 
\and Markus Wagner \affmark[3]\href{https://orcid.org/0000-0002-3124-0061}{\includegraphics[width=10pt,height=10pt]{orcid.pdf}}}
}

\authorrunning{Jonatas B. C. Chagas \and Markus Wagner} 

\institute{Jonatas B. C. Chagas \at
           \email{jonatas.chagas@iceb.ufop.br} 
           \and
           Markus Wagner \at
           \email{markus.wagner@adelaide.edu.au}
           \\ \\
           \affaddr{\affmark[*] Corresponding author} \\
           \affaddr{\affmark[1] \mbox{Departamento de Computação, Universidade Federal de Ouro Preto, Ouro Preto, Brazil}}\\
           \affaddr{\affmark[2] Departamento de Informática, Universidade Federal de Viçosa, Viçosa, Brazil}\\
           \affaddr{\affmark[3] School of Computer Science, The University of Adelaide, Adelaide, Australia}
}

\date{Received: date / Accepted: date}

\maketitle

\begin{abstract}
We tackle the Thief Orienteering Problem (ThOP), an academic multi-component problem that combines two classical combinatorial problems, namely the Knapsack Problem and the Orienteering Problem. In the ThOP, a thief has a time limit to steal items that distributed in a given set of cities. While traveling, the thief collects items by storing them in their knapsack, which in turn reduces the travel speed. The thief has as the objective to maximize the total profit of the stolen items. In this article, we present an approach that combines swarm-intelligence with a randomized packing heuristic. Our solution approach outperforms existing works on almost all the 432 benchmarking instances, with significant improvements.
\end{abstract}

\sloppy

\section{Introduction}
\label{sec:introduction}

Many studies devoted to solving classical combinatorial optimization problems are directly motivated by real-world problems, such as packing problems, scheduling problems, and vehicle routing problems. While already challenging, many real-world problems exhibit multi-component structures: they comprise a number of interacting combinatorial optimization problems. Multi-component problems are hard to solve as each component has the potential to influence the feasibility as well as the quality of the other components~\cite{bonyadi2019evolutionary}.

Among the studied multi-component problems, vehicle routing problems with loading constraints~\cite{iori2010routing} appear to be very frequently investigated. In these problems, tour are to be created for vehicles while constraints and aims of specific loading policies must be taken into account. One of these problems is the Traveling Thief Problem~(TTP), which combines two classic well-known and well-studied problems: the Knapsack Problem~(KP) and the Traveling Salesman Problem~(TSP). The TTP was proposed in 2013 by Bonyadi et al.~\cite{bonyadi2013travelling} in order to provide an academic abstraction of multi-component problems for the scientific community. 
In the TTP, a thief travels across all cities (TSP component) and steals items along the way (KP component). The stolen items are stored in a rented knapsack, for which the thief has to pay a time-dependent fee. 
The overall objective of the thief is to maximize the following: the total profit of all stolen items, minus the travel time multiplied with the renting rate. As the travel speed is inversely proportional to the total weight of the items in the knapsack, this non-constant cost sets the TTP apart from many vehicle routing problems, such as the environmental prize-collection problem~\cite{TRACHANATZI2020100712}.

In this article, we tackle the Thief Orienteering Problem (ThOP)~\cite{santos2018thief}, another academic multi-component problem. The ThOP has been proposed based on the TTP, but with different interactions and constraints in mind. It combines the Orienteering Problem (OP) and the Knapsack Problem (KP). In operational research, the OP has been the subject of many studies~\cite{GoLeVo87,GUNAWAN2016315}: a traveler starts at a predetermined location, travels through a region visiting checkpoints, and has to arrive at a control point within a given limited time. Because each checkpoint has a score, the participant's objective is to find a route that maximizes the total score, i.e., where the sum of scores of the visited checkpoints is the highest. In the ThOP context, the participant (i.e., the thief) does not score points by just visiting a checkpoint but has to steal valuable items located in them and carry the stolen items in their knapsack until the end of their robbery journey. As in the TTP, in the ThOP, the thief has a capacitated knapsack to carry the items. Moreover, as items are collected, the knapsack becomes heavier, and the speed of the thief decreases. There is no knapsack rental fee, and the thief only aims to find a route and a set of items that maximizes their total stolen profit.

Although the ThOP and the TTP are related, the ThOP appears to be more practical due to two key differences: in the ThOP (A) the participant does not have to travel through all cities, and (B) the interaction is not defined by the time-dependent rent for the knapsack, but by the time-constraint. Even though the relaxation of difference (A) may seem to be straightforward, handling this constraint is typically reflected in the design of heuristic~\cite{wagner2018case} and exact~\cite{wu2017exact,Neumann2019fptasPWT} algorithms, with only Chand and Wagner's~\cite{chand2016fast} Multiple Traveling Thieves Problem (MTTP) being an exception. Regarding (B), applications with routing time limit frequently arise in real-world scenarios, where there is insufficient time and/or capacity to visit/met all possible locations. Examples of this include cash logistics~\cite{journals/transci/OrlisBRD20}, 
urban crowd-sourcing~\cite{conf/hcomp/ChenCGMDC14}, and concert promotion~\cite{KIM2020106808}.

Santos and Chagas~\cite{santos2018thief} presented a Mixed Integer Non-Linear Programming formulation for the ThOP (although without results) and two simple heuristics. Afterwards, Fa{\^e}da and Santos~\cite{faeda2020genetic} proposed a genetic algorithm that performed better on most instances. We have also addressed the ThOP in a preceding article~\cite{chagas2020ants} with a two-phase approach based on Ant Colony Optimization (ACO) and a greedy heuristic to construct the route and the packing plan (stolen items) of the thief. Our ACO is able to find better solutions than other aforementioned algorithms for most instances because of its focus on creating efficient routes.

Here, we describe a number of improvements that we incorporated into our ACO algorithm, which made it more substantially effective with regard to the quality of the solutions found. In our computation experiments, we have investigated the importance of the parameters of our ACO algorithm considering these improvement changes. In addition, to make a fairer comparison among the other algorithms already proposed for the ThOP, we have also investigated the parameters of those algorithms and then evaluate their performances on a broad set of instances according to the results already presented in the literature.

In this article, we begin in Section~\ref{sec:problem_definition} with a formal description of the TTP, where we also present a mixed integer non-linear programming formulation. Subsequently, we present our new approach for solving the ThOP in Section~\ref{sec:max_min_ant_algorithm}.  Then, we report on our computational experiments in Section~\ref{sec:computational_experiments}. Lastly, we summarize our present work and outline future work.

\section{Problem definition}
\label{sec:problem_definition}

\subsection{Formal definition}

The Thief Orienteering Problem (ThOP) can be formally described as follows. There is a set $I = \{1, 2, \ldots, m\}$ of $m$ items and a set $C = \{1, 2, \ldots, n\}$ of $n$ cities. Each item $k \in I$ has a profit $p_k$ and weight $w_k$ associated. In addition, each item is associated with only a single city, but a city can have multiple items. Let us denoted by $I_{i}$ the set of items localized at city $i$. From the foregoing definition, $\bigcup_{\,i \in C} I_{i} = I$ and $I_{i}\,\bigcap\,I_{j} = \varnothing\;\; \forall i \in C,\;\forall j \in C \setminus \{i\}$. The items are scattered among all cities, except cities $1$ and $n$ $(I_{1} = I_{n} = \varnothing)$. Cities $1$ and $n$ are the cities where the thief starts and ends their journey. Let us denote by $A = \{(i, j),\; \forall i \in C \setminus \{n\},\; \forall j \in C \setminus \{1, i\}\}$ the set of arcs in which the thief can travel. For any pair of cities $i$ and $j$ with $(i, j) \in A$, the distance $d_{ij}$ between them is known. The thief can make a profit throughout their journey by stealing items and storing them in a knapsack with a limited capacity $W$. Moreover, the thief has a maximum time $T$ to complete their whole robbery journey. As stolen items are put into the capacitated knapsack, its weight increases and the thief's velocity decreases inversely proportional to the knapsack weight. Specifically, when the knapsack is empty, the thief can move with their highest velocity $v_{max}$. However, when the knapsack is completely full, the thief moves with the minimum speed $v_{min}$. In general terms, the thief can move with a speed $v = v_{max} - w \cdot (v_{max} - v_{min})\,/\, W$, where $w$ is the current weight of their knapsack. The objective of the ThOP is to find a path for the thief that starts from city $1$ and ends at city $n$, as well as a robbery plan, i.e., a set of items chosen from the cities visited that maximizes the total profit stolen, ensuring that the capacity of the knapsack $W$ is not exceeded and that the total traveling time is within the given time limit $T$.

We can represent any solution for the ThOP through a pair $\langle\pi, z\rangle$, where $\pi = \langle1, \ldots, n\rangle$ is a list of visited cities by the thief, and $z = \langle z_1, z_2, \ldots, z_m \rangle$ is a binary vector to represent the packing plan ($z_j = 1$ if item $j$ is collected, and $0$ otherwise) adopted by the thief throughout their robbery journey. Note that the first and last cities are fixed for any feasible solution. In addition, the number of cities visited may differ for different solutions. 

It is important to note that nothing prevents the thief from visiting a city more than once. In this scenario, the thief should collect all items of a city at once on the last visit from that city to minimize their travel time and consequently has more time to collect other items. One more aspect that should be pointed out is that a solution that visits some cities without collecting any items would only be advantageous if the distances between cities do not respect the triangular inequality because it may be convenient that the thief visits a city just to shorten their route. Nevertheless, as Santos and Chagas~\cite{santos2018thief} have defined the test problems for the ThOP -- which we have also used in this work -- in such a way that the distances between cities respect triangular inequality, it can be considered as an implicit optimization that any solution is formed by a list of cities $\pi$ without repetition.

\subsection{Mixed Integer Non-Linear Programming formulation}

In order to formally describe the ThOP through a mathematical formulation, we have proposed an alternative Mixed Integer Non-Linear Programming (MINLP) formulation to that proposed by Santos and Chagas~\cite{santos2018thief}. In contrast to Santos and Chagas' formulation, the following formulation uses a polynomial number of decision variables in terms of the number of cities and items. These decision variables are detailed below:

\begin{itemize}
    \setlength\itemsep{1mm}
   	\item { 
   	    $x_{ij}:$ binary variable that gets $1$ if the thief crosses arc $(i, j) \in A$, and $0$ otherwise.
	}
	\item {
	    $y_{i}:$ binary variable that gets $1$ if the thief visits city $i \in C$, and $0$ otherwise.
	}
	\item {
	    $z_{k}:$ binary variable that gets $1$ if the thief collects item $k \in I$, and $0$ otherwise.
	}
    \item {
   		$q_{i}:$ variable that reports the weight of the knapsack after leaving city $i \in C$.
   	}
   	\item {
   		$t_{i}:$ variable that informs the thief's arrival time at city $i \in C$.
   	}
\end{itemize}

With these variables, we can describe the following MINLP formulation for the ThOP.

\begin{equation} \label{eq:thop_model_obj}
    \max\;\;\displaystyle \sum_{k\,\in\,I}\;p_{k} \cdot z_{k}
\end{equation}%
\begin{align}
    \label{eq:thop_model_constr01}
    &\sum_{k \in I}\;w_{k} \cdot z_{k} \leq W \\
    \label{eq:thop_model_constr02}
    &y_{i} \geq z_{k} &i \in C, k \in I_{i} \\[1mm]
    \label{eq:thop_model_constr03}
    &y_{i} \leq \sum_{k \in I_{i}} z_{k} &i \in C \setminus \{1, n\} \\
    \label{eq:thop_model_constr04}
    &y_{1} = y_{n} = 1
\end{align}%
\begin{align} 
    \label{eq:thop_model_constr05}
    &\sum_{j:(i, j)\;\in\;A} x_{ij} = y_{i} &i \in C \setminus \{n\} \\
    \label{eq:thop_model_constr06}
    &\sum_{i:(i, j)\;\in\;A} x_{ij} = y_{j} &j \in C \setminus \{1\} \\
    \label{eq:thop_model_constr07}
    &q_{j} \geq \Bigg(q_{i} + \sum_{k \in I_{j}} w_{k} \cdot z_{k}\Bigg) \cdot x_{ij} &(i, j) \in A \\
    \label{eq:thop_model_constr08}
    &t_{j} \geq \Bigg(t_{i} + \frac{d_{ij}}{v_{max} - \nu \cdot q_{i}}\Bigg) \cdot x_{ij} &(i, j) \in A \\
    \label{eq:thop_model_constr09}
    &x_{ij} \in \{0, 1\} &(i, j) \in A \\[1mm]
    \label{eq:thop_model_constr10}
    &y_{i} \in \{0, 1\} &i \in C \\[1mm]
    \label{eq:thop_model_constr11}
	&z_{k} \in \{0, 1\} &k \in I \\[1mm]
	\label{eq:thop_model_constr12}
	&0 \leq q_{i} \leq W &i \in C \\[1mm]
	\label{eq:thop_model_constr13}
 	&0 \leq t_{i} \leq T &i \in C
\end{align}

The objective \eqref{eq:thop_model_obj} is to maximize the total profit of items collected. Constraint~\eqref{eq:thop_model_constr01} ensures that the total weight of items collected does not exceed the knapsack capacity. While constraints~\eqref{eq:thop_model_constr02} guarantee that the thief must visit a city to collect any item from it, constraints~\eqref{eq:thop_model_constr03} ensure that the thief does not visit cities where no items are selected. Note that constraints~\eqref{eq:thop_model_constr03} are not needed for the model to produce feasible solutions. However, these constraints strengthen the model by removing unprofitable route combinations. Constraint~\eqref{eq:thop_model_constr04} simply imposes that cities $1$ and $n$ have to be visited once they are, respectively, the fixed start and end points of any feasible route. Constraints~\eqref{eq:thop_model_constr05} and \eqref{eq:thop_model_constr06} guarantee route connectivity. Constraints~\eqref{eq:thop_model_constr07} and \eqref{eq:thop_model_constr08} guarantee that the knapsack weight and the route time is properly increasing along the route according to the items, which also avoid subcycles. Note that constraints~\eqref{eq:thop_model_constr07} and \eqref{eq:thop_model_constr08} are non-linear. Finally, constraints~\eqref{eq:thop_model_constr09}-\eqref{eq:thop_model_constr13} define the scope and domain of the decision variables. Note that, as constraints \eqref{eq:thop_model_constr12} ensure that the knapsack weight must be always less than knapsack capacity $W$ throughout the route, constraint~\eqref{eq:thop_model_constr01} could be removed. However, constraints \eqref{eq:thop_model_constr12} may be weaker for the purpose expressed by constraint~\eqref{eq:thop_model_constr01} due to the multiplication by the variable $x_{ij}$, which allows that weak fractional solutions to be considered during the resolution of the formulation.

It is worth mentioning that constraints~\eqref{eq:thop_model_constr07} can be linearized by rewriting them using Big-M constants as shown in~\eqref{eq:thop_model_constr07_line}. On the other hand, we cannot linearize constraints~\eqref{eq:thop_model_constr08} as they involve divisions and multiplications of decision variables. Nevertheless, we have also rewritten them using Big-M constants, as shown in~\eqref{eq:thop_model_constr08_line}, to remove their non-linear multiplications. In constraints~\eqref{eq:thop_model_constr07_line} and \eqref{eq:thop_model_constr08_line}, we have used different Big-M constants $M'_{j}$ and $M''_{ij}$, which should assume any sufficiently large number that is greater than or equal to $W + \sum_{i \in I_j} w_{i}$ and $T + d_{ij} / v_{min}$, respectively.

\begin{align}
    \label{eq:thop_model_constr07_line}
    &q_{j} \geq q_{i} + \sum_{k \in I_{j}} w_{k} \cdot y_{k} - M'_{j} \cdot \big(1 - x_{ij}\big) &(i, j) \in A \\
    \label{eq:thop_model_constr08_line}
    &t_{j} \geq t_{i} + \frac{d_{ij}}{v_{max} - \nu \cdot q_{i}} - M''_{ij} \cdot \big(1 - x_{ij}\big) &(i, j) \in A 
\end{align}

Although the foregoing mathematical formulation may be used for solving the ThOP from a mathematical solver, we have not considered it in our experiments due to its complexity. As constraints~\eqref{eq:thop_model_constr08_line} are still non-linear, they greatly increase the complexity of the formulation, making it impracticable to solve even the smallest-size instance defined in the literature for the ThOP~\cite{santos2018thief}. In fact, our formulation cannot find even reasonable bounds for the problem due to its non-linearity. Therefore, we have bet in a heuristic strategy for helping the thief in their robbery, leaving an improved mathematical formulation and exact algorithms for future investigation. As our MINLP formulation might be used as a starting point for other investigations, we have made it publicly available at \href{https://github.com/jonatasbcchagas/minlp_thop}{\textcolor{blue}{https://github.com/jonatasbcchagas/minlp\_thop}}\footnote{In addition to the code of our mathematical model, it also contains some experiments and results on smaller instances that we have created in order to solve the model and establish a benchmarking with exact results. However, the experiments have shown that even for instances with 15 cities and 1 item per city, our model cannot be solved, and it also is not able to find tight mathematical bounds within a reasonable computational time.}, which has been implemented using PySCIPOpt \cite{PySCIPOpt2016}, a Python interface for the SCIP Optimization Suite \cite{GamrathEtal2020ZR}.

\section{Problem-solving methodology}
\label{sec:max_min_ant_algorithm}

Throughout this section, we describe our solution approach, called ACO\texttt{++}, for solving the ThOP. Our ACO\texttt{++} is an improved version of the ACO algorithm previously presented in~\cite{chagas2020ants}. At the end of this section, we highlight the differences between both algorithms.

\subsection{The overall algorithm}

Our solution approach has been loosely based on Wagner's TTP approach~\cite{wagner2016stealing}, with Ant Colony Optimization (ACO)~\cite{dorigo1999ant} as the central component.

Following~\cite{wagner2016stealing}, we have used the ACO for determining the thief's route, while another algorithm for determining their packing plan for each route found by the ants. We have used the MAX-MIN ant system~\cite{stutzle2000max}, which limits all pheromones to an interval. In our implementation, we have used St{\"u}tzle's ACOTSP 1.0.3 framework\footnote{Publicly available online at \href{http://www.aco-metaheuristic.org/aco-code}{\textcolor{blue}{http://www.aco-metaheuristic.org/aco-code}}} for constructing the thief's route. The ACOTSP is an efficient framework that implements several ACO algorithms for the symmetric TSP. Most of the ACOTSP framework remains untouched in our approach, with only a few minimal modifications necessary to adapt it to the ThOP specifications. First, to construct the feasible routes for the thief from a given city to a given destination, and without the need to visit all cities, we have made an adaptation so that the ants built their routes until the thief's destination city, that is, city $n$ has been visited. Thus, the ants are able to build routes of varying sizes. 

Second, the pheromone trail updates are performed based on the quality of the TSP routes. Because the TSP's objective is to find the shortest possible route over all cities, the fitness of a given route is inversely proportional to its total distance. In contrast to this, in our ACOTSP adaptation, the fitness of each route is set based on the profit of the stolen items, because stolen items define the quality of ThOP solutions, which are defined by our proposed packing routine (to be described later): the fitness of a ThOP's route $\pi$ is inversely proportional to $UB + 1 - p(z)$, where $UB$ is an upper bound for the ThOP and $p(z)$ is the total profit of packing plan $z$. Thus, the fitness behaves similar to that of the TSP, and we do not need to modify the ACOTSP any further. To set the upper bound $UB$, we use the optimal solution for the KP version that allows the selection of fractions of items, which can be solved in $\mathcal{O}(m\log_2 m)$~\cite{toth1990knapsack}.

In Algorithm~\ref{alg:aco++thop}, we show the simplified overview of our ACO\texttt{++}. At the beginning (Line~\ref{alg:best_sol_init}), the best ThOP solution (route and packing plan) found by the algorithm is initialized as an empty solution. The algorithm performs its iterative cycle (Lines~\ref{alg:stopping_criterion_begin} to \ref{alg:stopping_criterion_end}) as long as the stopping criterion is not met. At Line~\ref{alg:construct_routes}, each ant constructs a route for the thief, and then packing plans are created  (Line~\ref{alg:for_each_route_begin} and~\ref{alg:packing_plan}). The ACOTSP framework allows us to apply several classic local search heuristics: $2\text{-}opt$, $2.5\text{-}opt$, and $3\text{-}opt$~\cite{aarts2003local}. If any local search is enabled in our algorithm (Line~\ref{alg:local_search_enabled}), that local search procedure is performed on each route $\pi$, thus generating routes $\pi'$ (Line~\ref{alg:local_search}), which may be better than $\pi$ regarding the distance costs. In the next step, a packing plan $z'$ is created from $\pi'$ (Line~\ref{alg:new_packing_plan}). If $z'$ is better than $z$ (Line~\ref{alg:if_first_replace}), $\pi$ and $z$ are replaced by $\pi'$ and $z'$ (Line~\ref{alg:first_replace}). At Lines~\ref{alg:second_replace_begin} to \ref{alg:second_replace_end}, we update the best solution. Note that, to achieve more efficient routes, we remove from $\pi$ all those cities from which no items are stolen (Line \ref{alg:clear_route}). As we have stated before, this is only true as all ThOP instances use distances that preserve the triangular inequality. After every route has been considered, the pheromones are updated according to the quality of the ThOP solutions (Line \ref{alg:update_pheromone}). In the end, the best solution found is returned.

\begin{algorithm}[!ht]
\makeatletter
\newcommand{\algorithmfootnote}[2][\footnotesize]{%
  \let\old@algocf@finish\@algocf@finish
  \def\@algocf@finish{\old@algocf@finish
    \leavevmode\rlap{\begin{minipage}{\linewidth}
    #1#2
    \end{minipage}}%
  }%
}
\normalsize
\DontPrintSemicolon
\SetKwData{Left}{left}
\SetKwData{Up}{up}
\SetKwFunction{FindCompress}{FindCompress}
\SetKwInOut{Input}{input}
\SetKwInOut{Output}{output}
$\pi^{best} \gets \varnothing, z^{best} \gets \varnothing$ \label{alg:best_sol_init} \\
\Repeat{\upshape stopping condition is fulfilled} { \label{alg:stopping_criterion_begin}
    $\Pi \gets$ construct routes using ants \label{alg:construct_routes} \\
    \ForEach{\upshape route $\pi \in \Pi$} { \label{alg:for_each_route_begin}
        $z \gets$ construct a packing plan from $\pi$ \label{alg:packing_plan} \\
        \If{\upshape local search procedure is activated} { \label{alg:local_search_enabled}
            $\pi' \gets$ perform a local search procedure on route $\pi$ \label{alg:local_search} \\
            $z' \gets$ construct a packing plan from $\pi'$ \label{alg:new_packing_plan} \\
            \If{\upshape profit of $z'$ is higher than profit of $z$} { \label{alg:if_first_replace}
                $\pi \gets \pi'$, $z \gets z'$ \label{alg:first_replace}
            }
        }
        \If{\upshape profit of $z$ is higher than profit of $z^{best}$} { \label{alg:second_replace_begin}
            $\pi^{best} \gets \zeta(\pi$), $z^{best} \gets z$ \label{alg:clear_route}
        } \label{alg:second_replace_end}
    } \label{alg:for_each_route_end}
    update ACO statistics and pheromone trail \label{alg:update_pheromone} \\
} \label{alg:stopping_criterion_end}
\Return $\pi^{best}$, $z^{best}$
\caption{ACO\texttt{++} algorithm for the ThOP}
\label{alg:aco++thop}
\algorithmfootnote{$\zeta(\pi)$ removes from $\pi$ all cities where no items are stolen according to the packing plan $z$.}
\end{algorithm}

\subsection{Randomized packing heuristic}

To construct a packing plan in our proposed ACO\texttt{++} for a given route $\pi$, we use our previously developed, randomized heuristic for solving the ThOP~\cite{chagas2020ants}:\footnote{As stated by Polyakovskiy and Neumann~\cite{polyakovskiy2015packing},  determining the optimal packing plan is $\mathcal{NP}$-hard, even when the route of the thief is kept fixed.} in brief, the randomization varies the relative influence of weights, profits, and distances in the (originally deterministic) heuristic \textsc{PackIterative}, which is an efficient packing algorithm developed for the TTP~\cite{faulkner2015approximate}. This non-deterministic strategy lets us explore the packing plan space even for a fixed route, which can lead to overall better configurations. This was also necessitated by the observation that the ants would often find near-identical routes during an optimization run.

\subsection{Differences between the ACO\texttt{++} and ACO approaches}

Compared to the ACO described in~\cite{chagas2020ants}, we have incorporated two new features into our ACO\texttt{++}. These features are described in the following:

\begin{enumerate}
    \item The ants of ACO\texttt{++} construct routes that do not necessarily visit all cities, while in our previous ACO the ants always construct complete TSP tours, i.e., all cities are visited. Note that, although both algorithms remove from the route all cities in which no item is selected, there is now a higher consistency with respect to the ThOP's definition, because ants do not have to visit all cities. 
    Note that this allows ants to construct routes that might not be easily constructed from our previous ACO. In addition, when a route visits fewer cities fewer items are available to be collected from that route, which reduces the search space to find a packing plan, thus making our packing routines more computationally efficient. It is also because of this last point, that we had to adapt our randomized packing heuristic algorithm to consider only the items that can be selected from each constructed route. However, its core idea remains unchanged.
    \item There is now the possibility to apply different local searches on each route constructed by the ants in our ACO\texttt{++}. While this results in a TSP-bias toward shorter routes, finding shorter routes may help the thief to better plan their robbery journey.
\end{enumerate}

\section{Computational study}
\label{sec:computational_experiments}

In the following, we present the experiments performed to study the performance of the proposed algorithm against other algorithms proposed for the ThOP~\cite{santos2018thief,faeda2020genetic,chagas2020ants}. As the computational budget of all ThOP algorithms are based on wallclock time, in order to enable a fair comparison, we have rerun all ThOP codes, except for Fa{\^e}da and Santos~\cite{faeda2020genetic}'s algorithm because we have not had access to their code. In our experiments, we have used a machine with Intel(R) Xeon(R) CPU X5650 @ 2.67GHz, running CentOS 7.4.

Our algorithm has been implemented based on Thomas Stützle’s ACOTSP 1.0.3 framework, which has been in C programming language. All raw results and solutions (tours and packing plans), as well as the code, are publicly available at \href{https://github.com/jonatasbcchagas/acoplusplus_thop}{\textcolor{blue}{https://github.com/jonatasbcchagas/acoplusplus\_thop}}. 

\subsection{Benchmarking instances}
\label{section:benchmarking_instances}

To evaluate the different ThOP approached, we use all 432 ThOP instances from~\cite{santos2018thief}. These have been created based on the TTP instances~\cite{polyakovskiy2014comprehensive} by removing the items in city $n$ and by adding a maximum travel time. The instances have the following characteristics:

\begin{itemize}
    \item[$\bullet$] {
        numbers of cities: 51, 107, 280, and 1000
        (TSP instances: {\it eil51}, {\it pr107}, {\it a280}, {\it dsj1000});
    }
    \item[$\bullet$] {
        numbers of items per city: {\it 01}, {\it 03}, {\it 05}, and {\it 10};
    }
    \item[$\bullet$] {
        sizes of knapsacks: {\it 01}, {\it 05} and {\it 10} times the size of the smallest knapsack;
    }
    \item[$\bullet$] {
        types of knapsacks: values and weights of the items are either uncorrelated ({\it unc}),  uncorrelated with similar weights ({\it usw}), or bounded and strongly correlated ({\it bsc});
    }
    \item[$\bullet$] {
        maximum travel times: {\it 01}, {\it 02}, and {\it 03} classes. These values refer to 50\%, 75\%, and 100\% of instance-specific references times defined in the original ThOP paper~\cite{santos2018thief}.
    }
\end{itemize}

In the remainder of this article, each instance will be identified as {\tt XXX\_YY\_ZZZ\_WW\_TT}, where {\tt XXX}, {\tt YY}, {\tt ZZZ}, {\tt WW} and {\tt TT} indicate the different characteristics of the instance at hand. For example, {\tt pr107\_05\_bsc\_01\_01} identifies the instance with 107 cities (TSP instance {\it pr107}), 5 items per city with their weights and values bounded and strongly correlated with each other, and the smallest knapsack and time limit defined.

\subsection{Parameter tuning}
\label{sec:parameter_tuning}

In the following, we tune a variety of ACOTSP parameters: \textit{ants} defines the number of ants; \textit{alpha} controls the relative importance of pheromone values in the construction of routes; \textit{beta} defines the influence of distances between cities; \textit{rho} is the evaporation rate of the pheromones; and \textit{localsearch} controls whether and what local search procedure to apply. Moreover, we vary  the number of attempts of our randomized packing heuristic \textit{ptries}. To stop the algorithm, as in previous work on the ThOP, we limit the execution time to $\lceil0.1m\rceil$ seconds, which is determined based on the number of items $m$ of each particular instance.

To find well-performing configurations, we have followed the same tuning experiments used in~\cite{chagas2020ants}, i.e., we have used the Irace package~\cite{lopez2016irace} for the automatic configuration of algorithms~\cite{birattari2010f}, in order to determine the influence of parameter values across different instance sets. We have divided all 432 instances into 48 groups and then executed Irace on each of them to  achieve tuned configurations that (1) perform well and (2) that can be analyzed to learn about the problem domain. Each instance group is identified as {\tt XXX\_YY\_ZZZ}: {\tt XXX} denotes the TSP base group, {\tt YY} the number of items per city and {\tt ZZZ} the knapsack type. Each group {\tt XXX\_YY\_ZZZ} contains all nine instances defined with different knapsack sizes and maximum travel time.

In Table~\ref{table:parameter_values} we show the parameters as well as their ranges; the ranges were determined in preliminary experiments. We have used Irace with its default settings, except for the parameter \textit{maxExperiments}, which we have set to 5000.

\begin{table}
\centering
\small
\caption{Parameter values considered during the tuning experiments.}
\setlength{\tabcolsep}{15pt}
\begin{tabular*}{\hsize}{@{}@{\extracolsep{\fill}}cc@{}}
\toprule
\multicolumn{1}{c}{Parameter} & \multicolumn{1}{c}{Investigated values} \\ 
\midrule
ants & $\{10, 20, 50, 100, 200, 500, 1000\}$ \\ 
alpha & $\{0.00, 0.01, 0.02, \ldots, 10.00\}$ \\ 
beta & $\{0.00, 0.01, 0.02, \ldots, 10.00\}$ \\ 
rho & $\{0.00, 0.01, 0.02, \ldots, 1.00\}$ \\ 
ptries & $\{1, 2, 3, 4, 5\}$ \\
localsearch &  $\{$no local search, $2\text{-}opt$, $2.5\text{-}opt$, $3\text{-}opt\}$ \\ 
\bottomrule
\end{tabular*}
\label{table:parameter_values}
\end{table}

\begin{figure*}[!ht]
\centering
 	\setcounter{subfigure}{0}
 	\subfloat {
      		\includegraphics[width=0.46\linewidth]{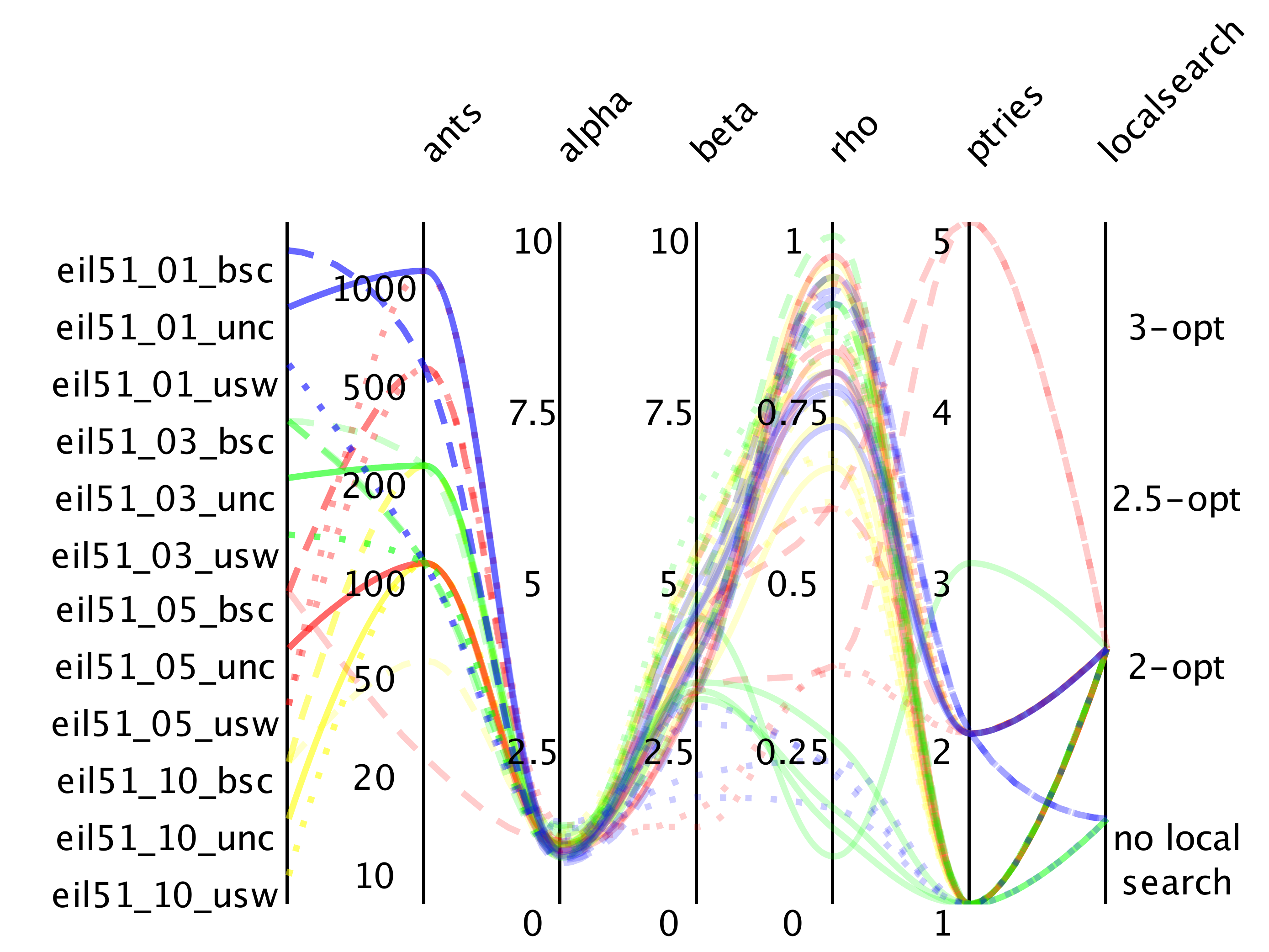}
 	}%
 	\quad
 	\subfloat {
      		\includegraphics[width=0.46\linewidth]{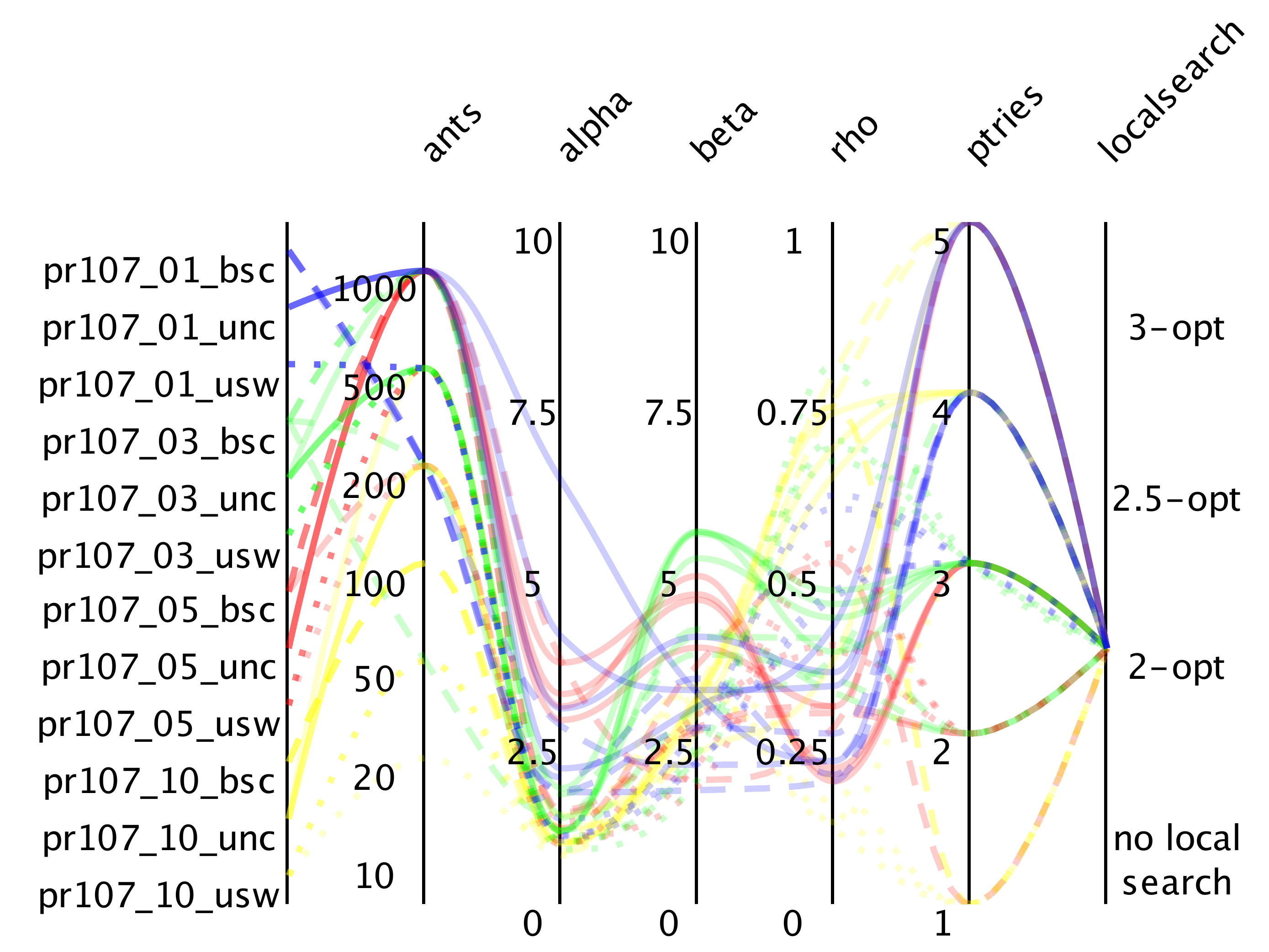}
 	}%
 	
 	\subfloat {
      		\includegraphics[width=0.46\linewidth]{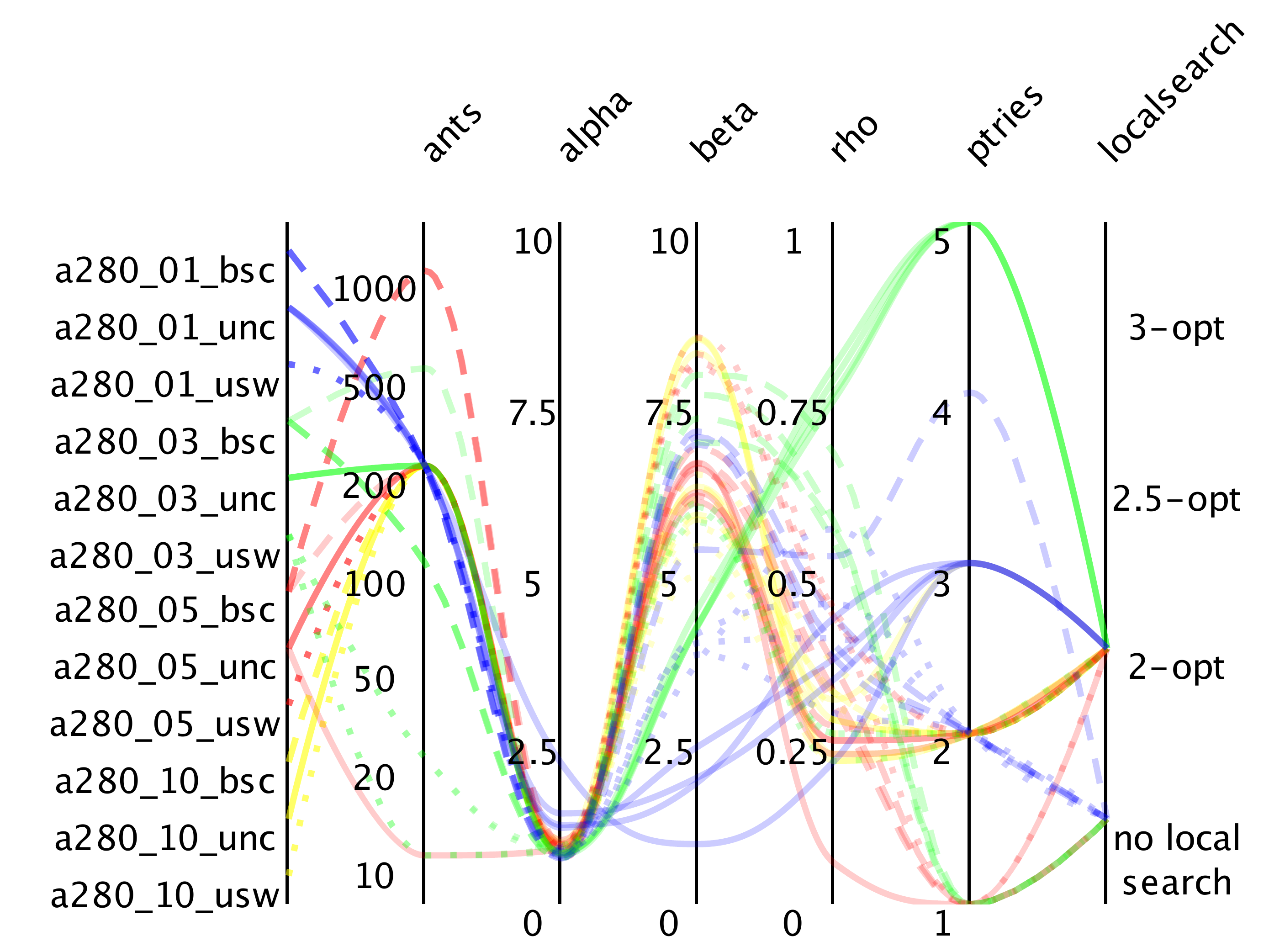}
 	}%
 	\quad
 	\subfloat {
      		\includegraphics[width=0.46\linewidth]{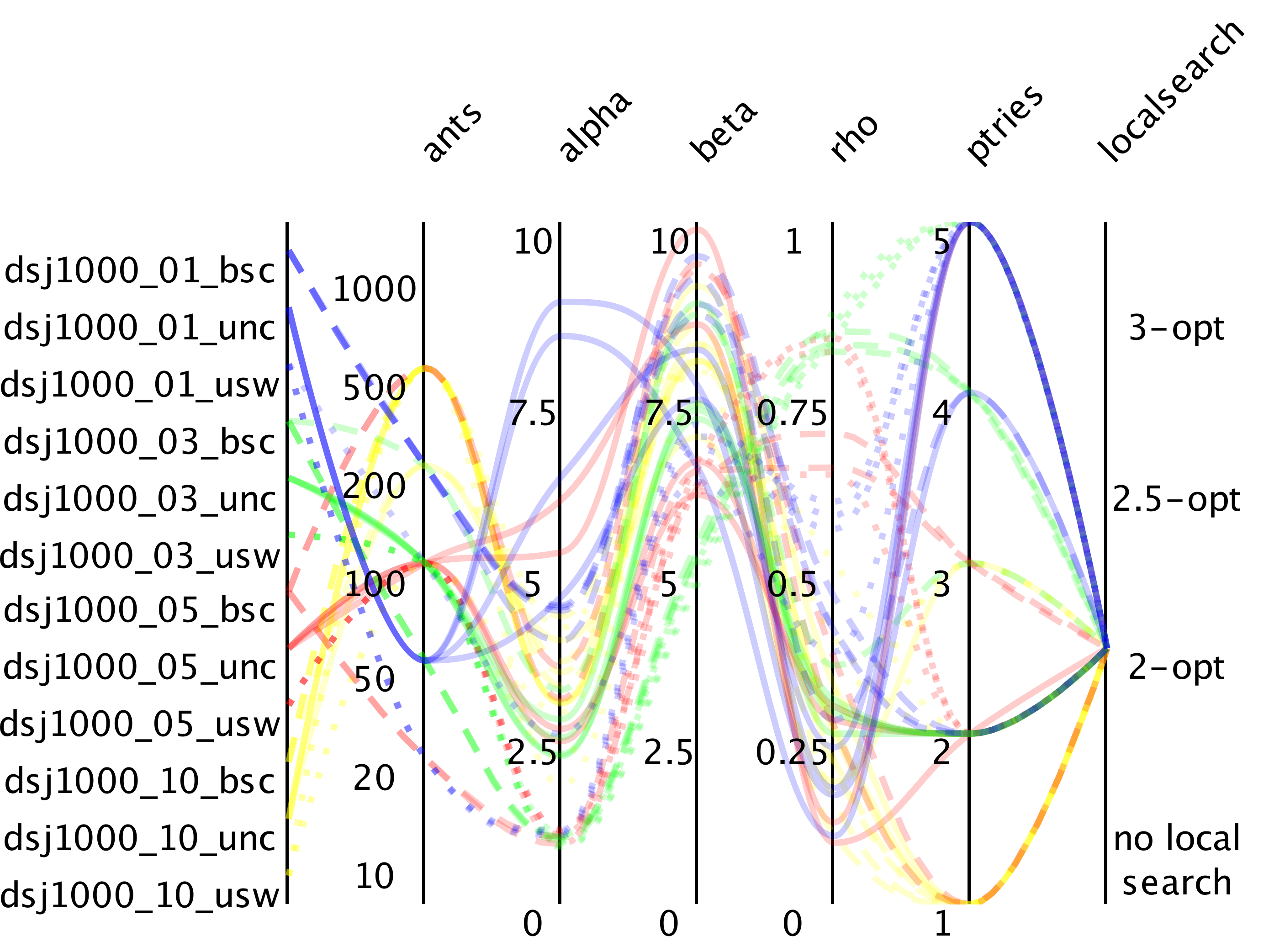}
 	}%
\caption{Irace results for the 48 instance groups. Dashed, solid, and dotted lines are used, respectively, to emphasize the groups of instances with items where their weights and values are uncorrelated ({\it unc}), uncorrelated with similar weights ({\it usw}), and bounded and strongly correlated ({\it bsc}). Blue, green, red, and yellow lines represent, respectively,  groups of instances with 1, 3, 5, and 10 items-per-city.}
\label{fig:irace_results}
\end{figure*}

In Figure \ref{fig:irace_results}, we show for each group of instances the configurations returned by Irace. Because Irace can return more than one configuration, we can sometimes see several configurations originating from the same instance (shown in the left-most columns). Each axis stands for a parameter and each parameter configuration is described by a line that intersects each parallel axis in its corresponding value. We can see which parameter values have been most selected among all tuning experiments by looking for ``concentrations'' of lines. We use different styles and colors to emphasize the results obtained for each individual group. All logfiles for these experiments can be found at the GitHub repository along with our code.

We can make several observations. First, we notice that the number of ants is typically between 50 and 200. The importance of the pheromone trail ($\alpha$) is typically low, especially for the groups of instances that consider the TSP bases {\it eil51}, and {\it a280}. In turn, the importance of distances between cities ($\beta$) varies depending on the underlying TSP instance. This is to be expected, because the underlying TSP instances are different in nature and not normalized, hence requiring different values of beta. The evaporation rate of the pheromone trail has had a behavior more spread, although it seems to have a compensation correlation between the parameter \textit{beta}: the higher the influence of distances between cities, the lower the evaporation rate of the pheromone trail. We can also observe that nearly all tuned configurations require the multiple invocation of our randomized packing heuristic, with the number of packing attempts widely spread among each other. Regarding the application of local searches on routes, one can note that for most groups of instances, the use of 2-opt moves produces better ThOP solutions. However, some configurations do not include the use of any local search. Note that there are no configurations that indicate the use of 2.5-opt and 3-opt moves. Potentially, this is because high-quality TSP routes do not necessarily result in high-quality ThOP routes. Therefore, there may be no need to use local searches with larger neighborhood moves based solely on route distances as an improvement phase for ThOP routes.

\subsection{Comparison of ThOP solution approaches}

We compare the quality of the solutions obtained by ACO\texttt{++} with the quality of the solutions obtained by other algorithms (ILS~\cite{santos2018thief}, BRKGA~\cite{santos2018thief}, GA~\cite{faeda2020genetic}, ACO~\cite{chagas2020ants}) already proposed for the ThOP. To enable a fair comparison, we also tuned the parameters of the BRKGA and ACO following the same process for ACO\texttt{++}, i.e., we have individually executed the Irace for each 48 different groups of instances. The ILS algorithm has no parameters to be tuned~\cite{santos2018thief}, while for the GA, we have not investigated its parameters because we have not had access to its code. Therefore, for the GA, we have made our analysis based on the results reported in~\cite{faeda2020genetic}.

Because all algorithms are randomized, we have performed 30 independent runs per instance. Each run has been executed with the parameter values with the best mean performance among those returned by Irace. ILS, BRKGA, and ACO codes, as well as their tuned configurations, raw results and solutions found, are also available at the GitHub link along with our ACO\texttt{++}.

\begin{figure*}[!ht]
\centering
 	\setcounter{subfigure}{0}
 	\subfloat[ILS] {
      		\includegraphics[width=0.32\linewidth]{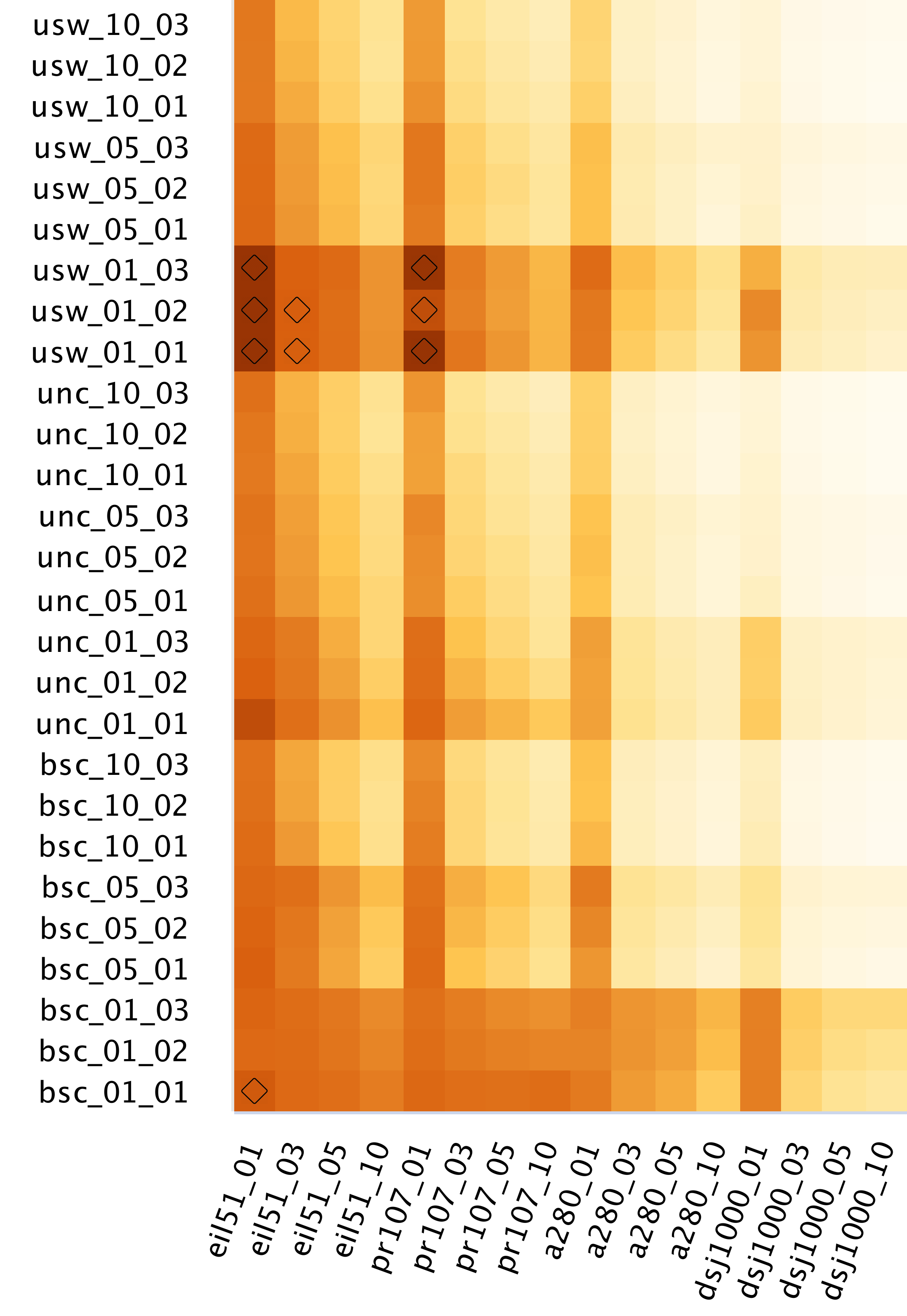}
 	}%
 	\subfloat[BRKGA] {
      		\includegraphics[width=0.32\linewidth]{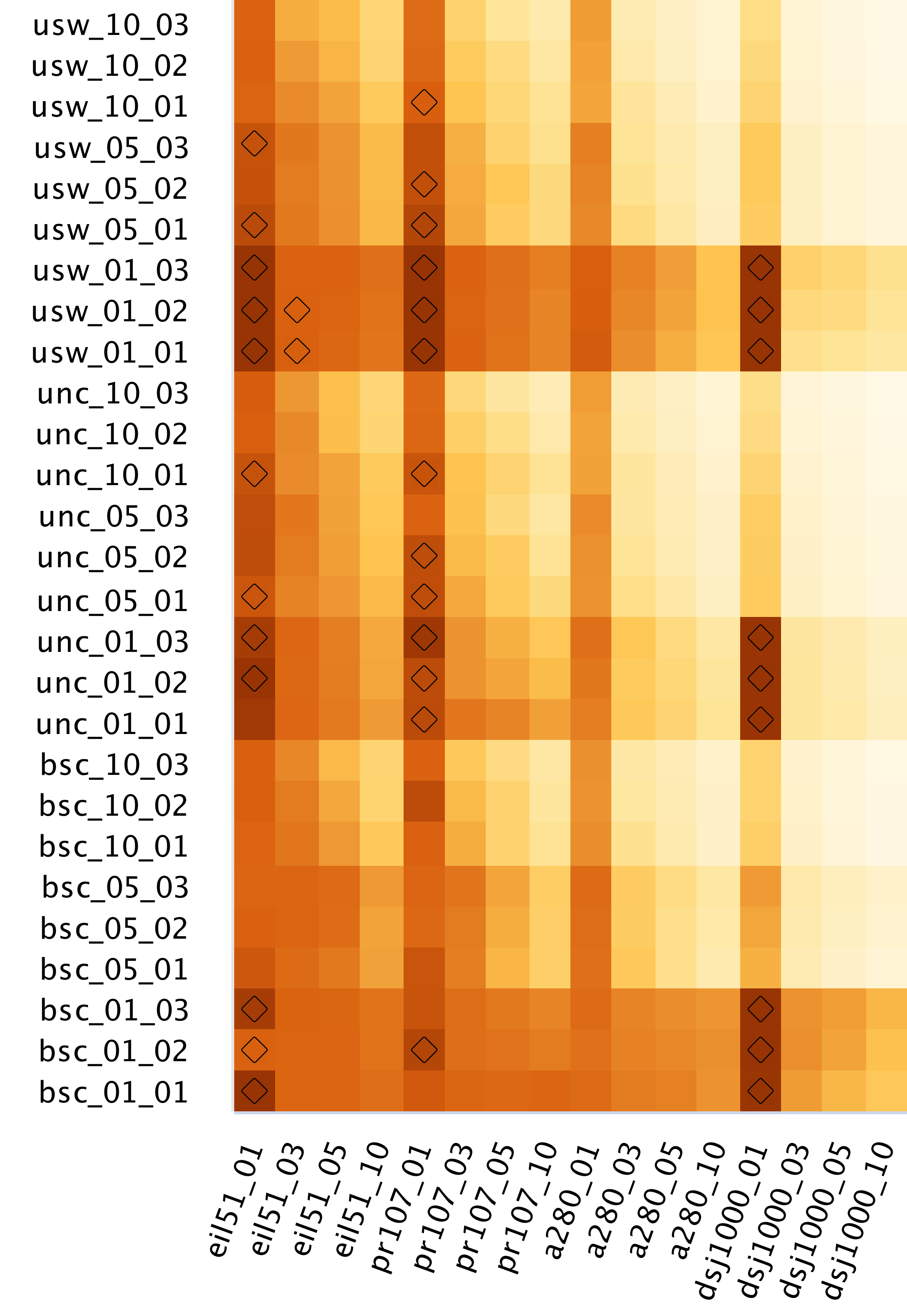}
 	}%
 	\subfloat[GA] {
      		\includegraphics[width=0.32\linewidth]{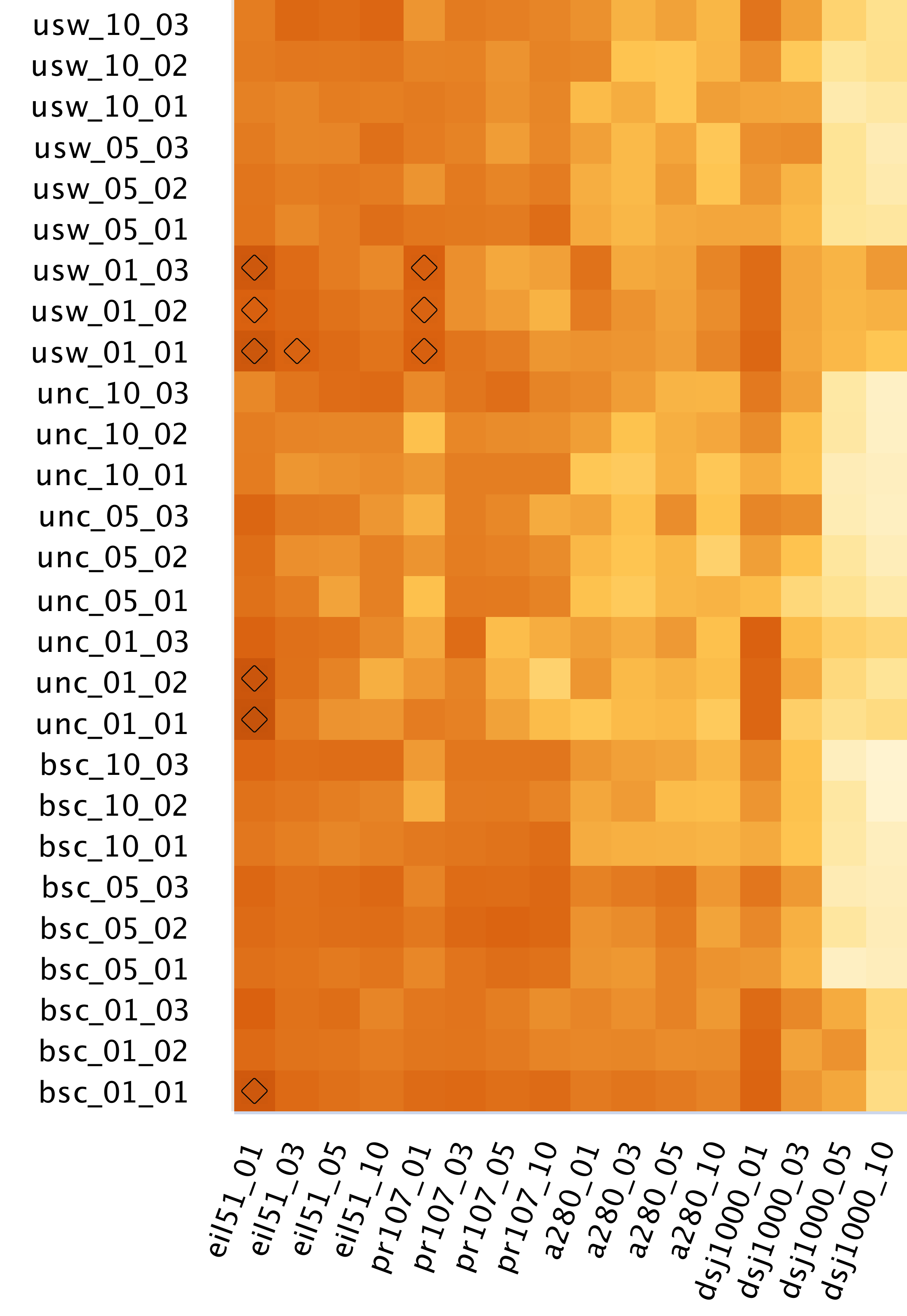}
 	}%
 	
 	\subfloat[ACO] {
      		\includegraphics[width=0.32\linewidth]{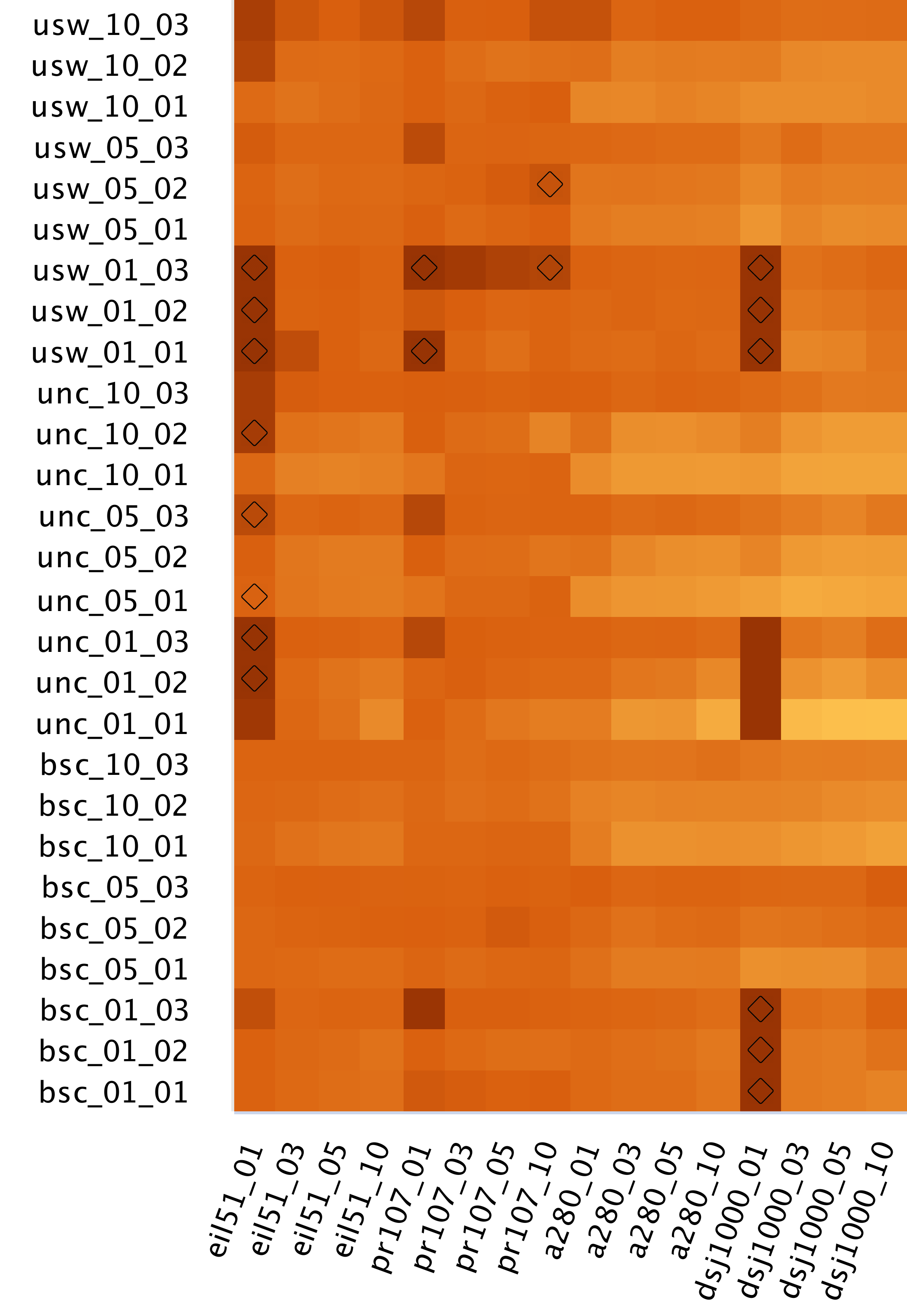}
 	}%
 	\qquad \qquad
 	\subfloat[ACO\texttt{++}] {
      		\includegraphics[width=0.32\linewidth]{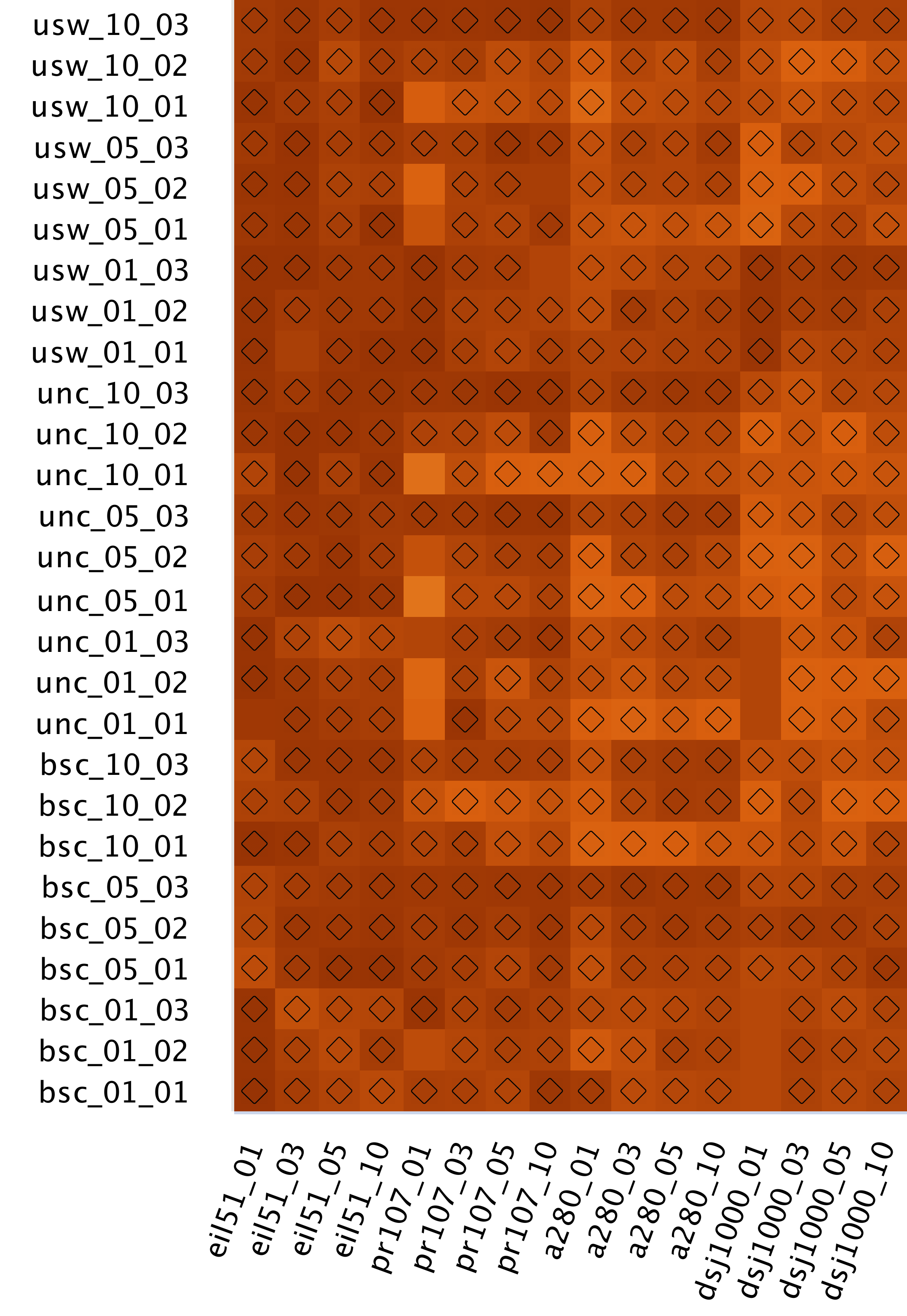}
 	}%
 	
 	\subfloat {
      		\includegraphics[width=0.85\linewidth]{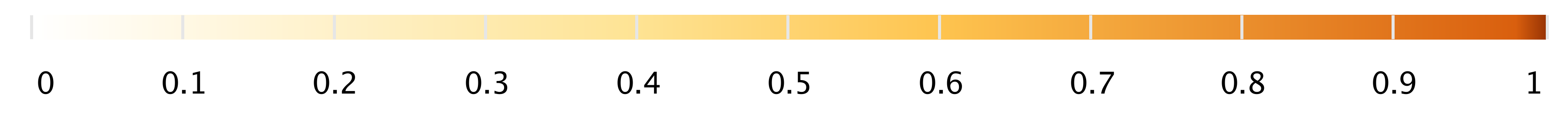}
 	}%
\caption{Approximation ratio of the solution approaches. Diamond symbols highlight in which the instances each algorithm has found the best solutions.}
\label{fig:approximation_ratio}
\end{figure*}

In the first analysis, we compare the performance of the solutions obtained by measuring for each instance the achieved approximation ratio: for each instance and algorithm, we take the average objective value obtained considering the independent runs of that algorithm and compute the ratio between that average objective value and the best objective value found among all algorithms. Note that the higher the approximation ratio, the higher the average performance of that particular algorithm. In Figure~\ref{fig:approximation_ratio}, we plot for every instance and algorithm the approximation ratio as a heatmap in order to highlight larger differences. Moreover, we use diamond symbols to highlight the instances for which each algorithm has found the best known solutions.

From Figure \ref{fig:approximation_ratio}, we can make several observations. As stated by Santos and Chagas~\cite{santos2018thief}, we can also confirm that their BRKGA has outperformed their ILS for most instances, with higher prominence on the larger-size instances. In addition, one can note that their algorithms perform better for instances that involve only one item per city. Regarding the best-known solutions, we can see that their algorithms have not been able to find many of them. The GA proposed by Fa{\^e}da and Santos~\cite{faeda2020genetic} has outperformed, in general, both BRKGA and ILS solution approaches. Although BRKGA has found more best-known solutions, the GA has a more uniform behavior regarding the dimensions of the instances. Note that our previous ACO algorithm~\cite{chagas2020ants} has reached a better approximation ratio for almost all instances when compared to GA and also to ILS and BRKGA. In turn, our current ACO\texttt{++} algorithm has presented a better or equal performance regarding the other algorithms for almost all instances. Similarly, it has typically found the best  solutions for most of the instances.

In order to compare each pair of algorithms as to the best solutions found by them, we show in Table~\ref{table:bks_pairwise} the percentage of the number of instances in which every algorithm found better or equal quality solutions than another algorithm. The results shown in this table corroborate with those shown in Figure~\ref{fig:approximation_ratio}. In addition to showing that the ACO\texttt{++} algorithm outperformed all other algorithms by more than 96\% of the total of instances, we can also see that our previous ACO also is more efficient than ILS, BRKGA, and GA by over 88\% of instances. In turn, GA is more efficient than ILS and BRKGA, and BRKGA overcomes ILS.

\begin{table}[!ht]
\centering
\small
\caption{Percentage of the number of instances in which algorithm $i$ found better or equal quality solutions than algorithm $j$.}
\setlength{\tabcolsep}{15pt}
\begin{tabular*}{\hsize}{@{}@{\extracolsep{\fill}}cccccc@{}}
\toprule
$i \downarrow \quad j \rightarrow$ & ILS & BRKGA & GA & ACO & ACO\texttt{++} \\ 
\midrule
ILS & - & 3.01\% & 20.37\% & 4.63\% & 2.55\% \\ 
BRKGA & 99.54\% & - & 37.50\% & 14.58\% & 8.56\% \\ 
GA & 81.71\% & 64.58\% & - & 2.78\% & 2.31\% \\ 
ACO & 96.99\% & 88.89\% & 98.61\% & - & 5.32\% \\ 
ACO\texttt{++} & 99.54\% & 96.06\% & 99.54\% & 98.15\% & - \\
\bottomrule
\end{tabular*}
\label{table:bks_pairwise}
\end{table}

As both algorithms based on ACO metaheuristics have had the best and most similar performances, we statistically compare the quality of their solutions using the Wilcoxon signed-rank test. At a significance level of 5\%, the performance of ACO\texttt{++} has been statistically worse than ACO in only 11 instances, in 12 instances there is no difference between the performance of both algorithms, while in 409 instances (about 95\% of total) ACO\texttt{++} has been better than ACO.

\begin{figure*}[!ht]
\captionsetup[subfigure]{labelformat=empty}
\centering
 	\subfloat[profit = 68653 \quad distance = 429] {
      		\includegraphics[width=0.44\linewidth]{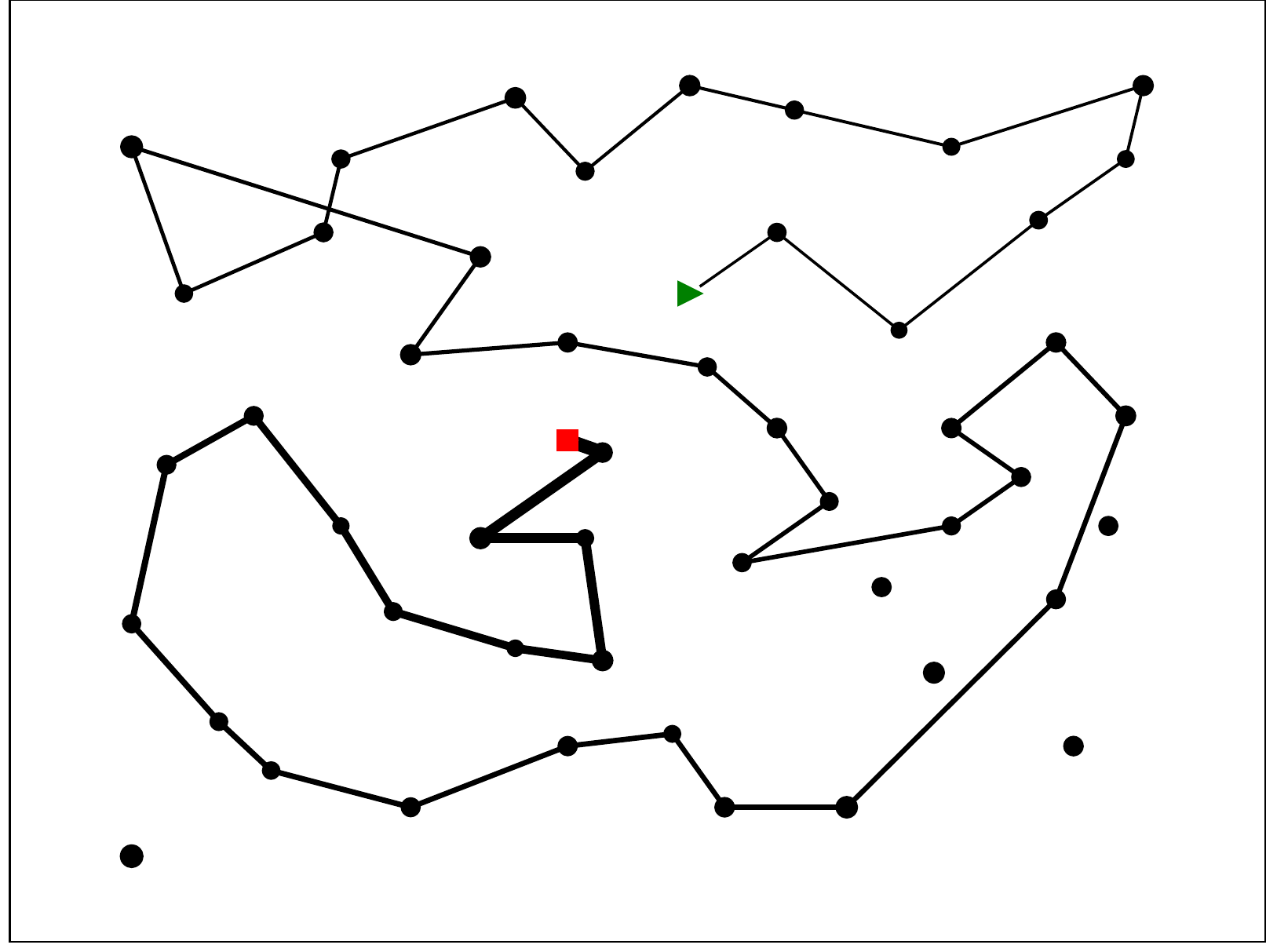}
 	}%
 	\subfloat[profit = 70830 \quad distance = 340] {
      		\includegraphics[width=0.44\linewidth]{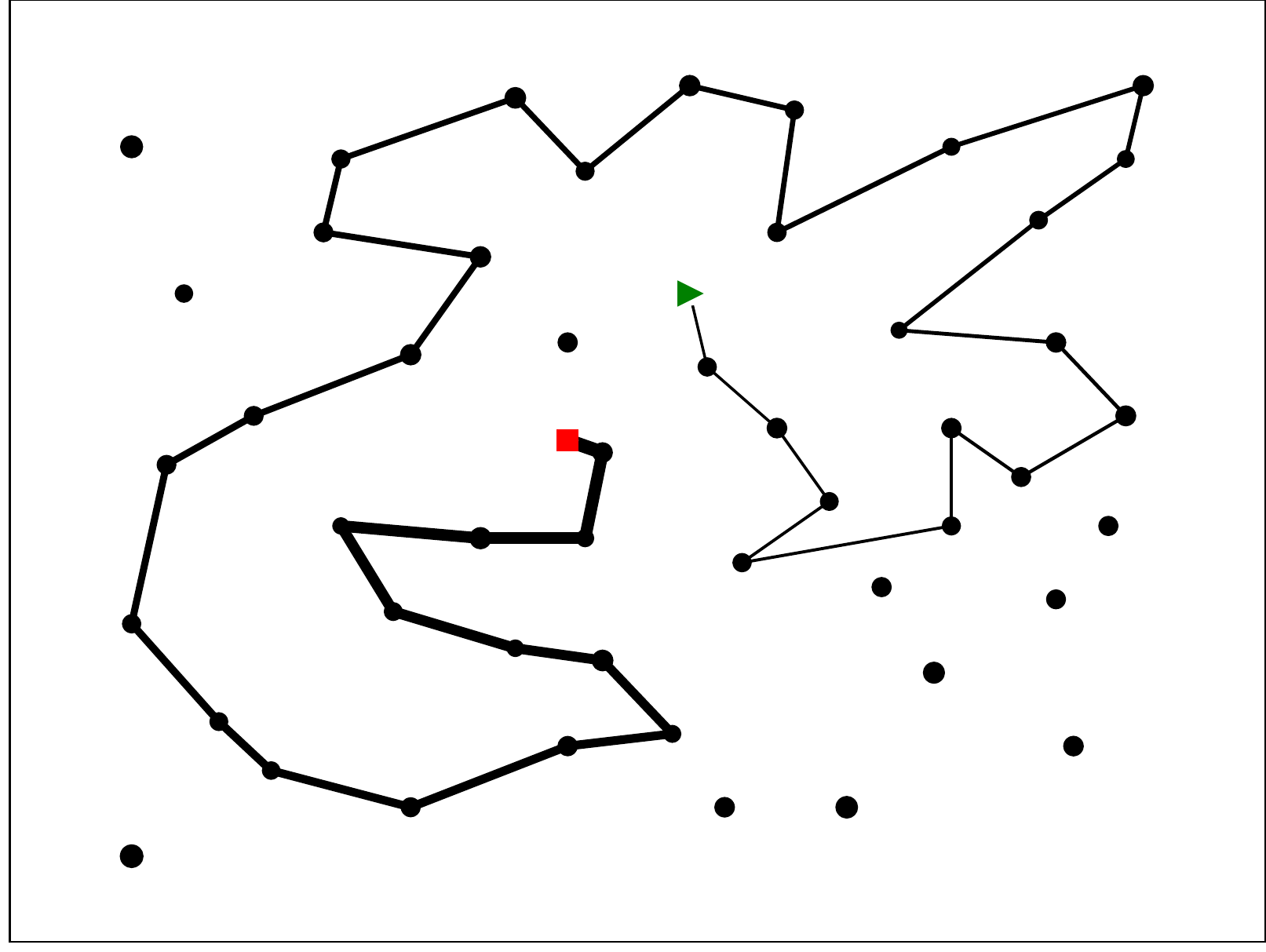}
 	}%
 	
 	\subfloat[profit = 180879 \quad distance = 34258] {
      		\includegraphics[width=0.44\linewidth]{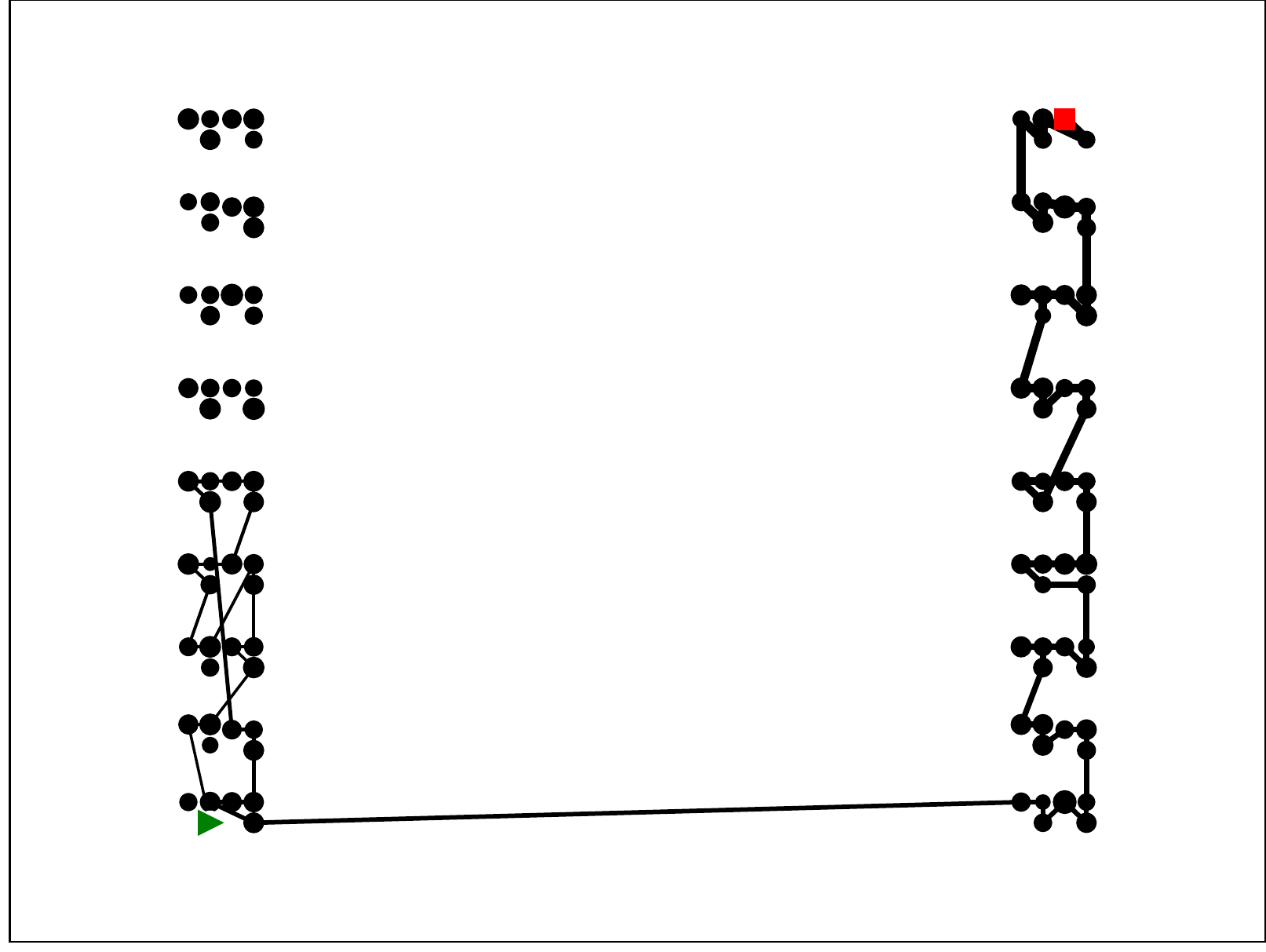}
 	}%
 	\subfloat[profit = 195716 \quad distance = 33796] {
      		\includegraphics[width=0.44\linewidth]{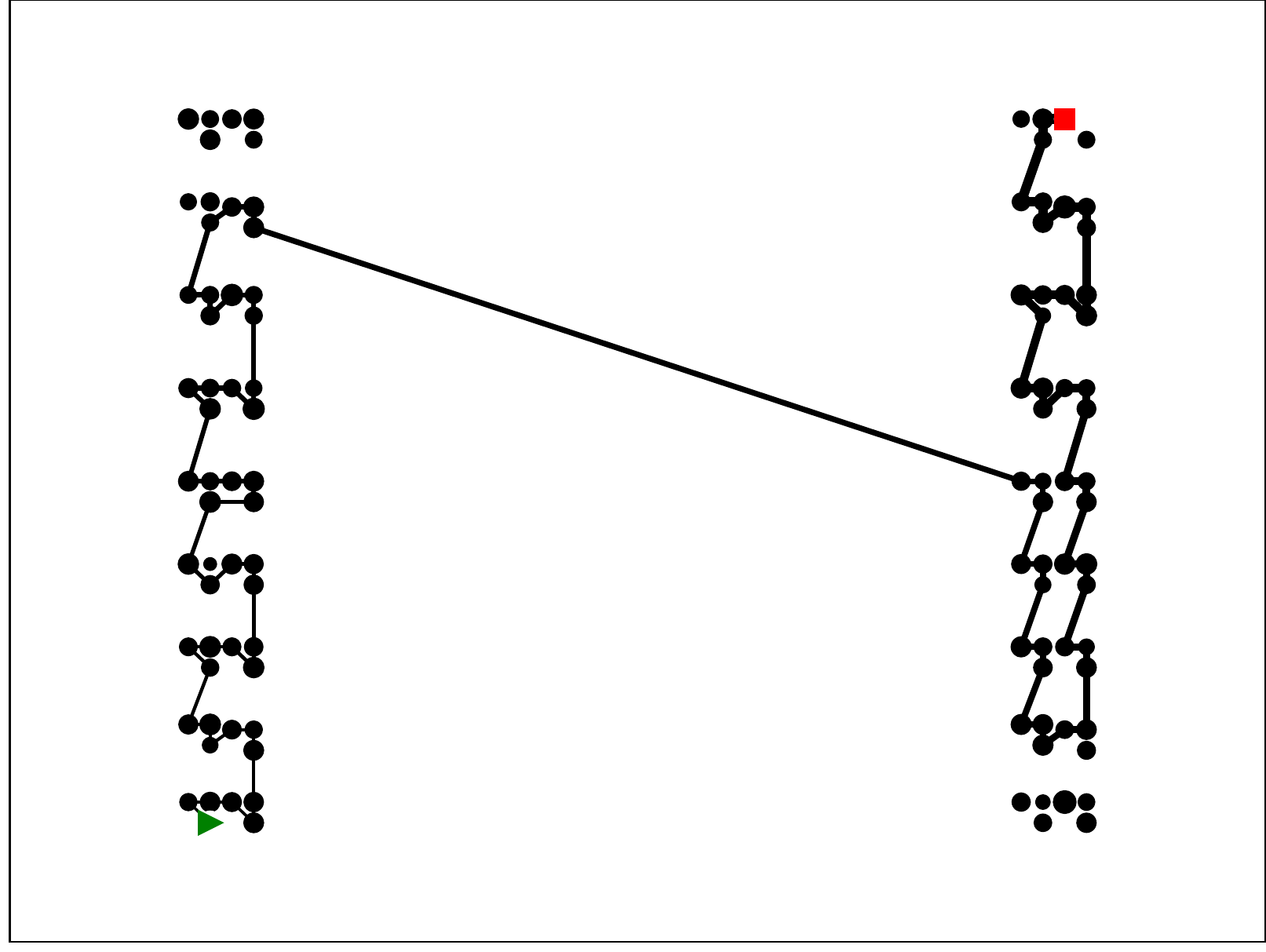}
 	}%
 	
 	\subfloat[profit = 422948 \quad distance = 2640] {
      		\includegraphics[width=0.44\linewidth]{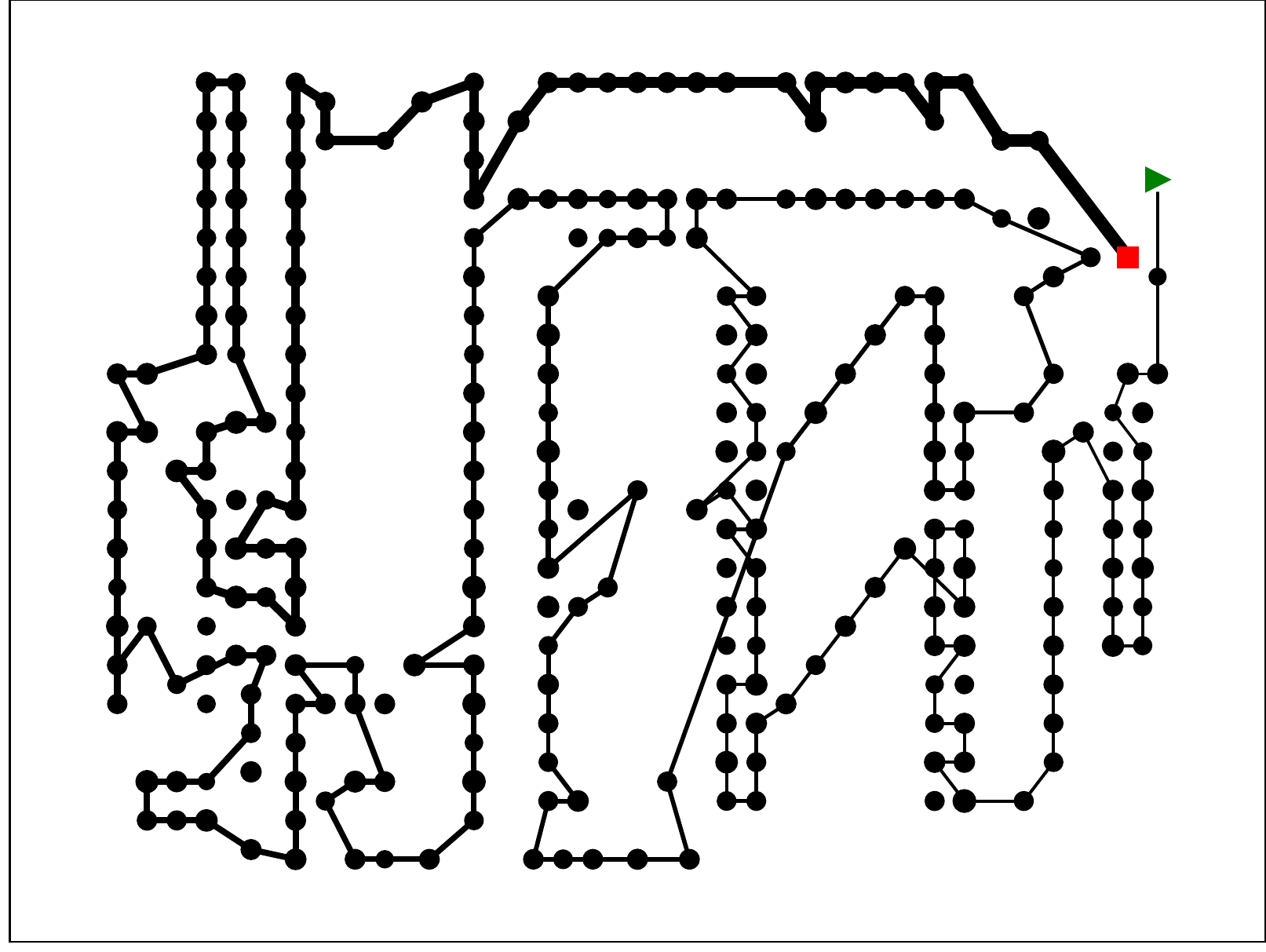}
 	}%
 	\subfloat[profit = 441530 \quad distance = 2348] {
      		\includegraphics[width=0.44\linewidth]{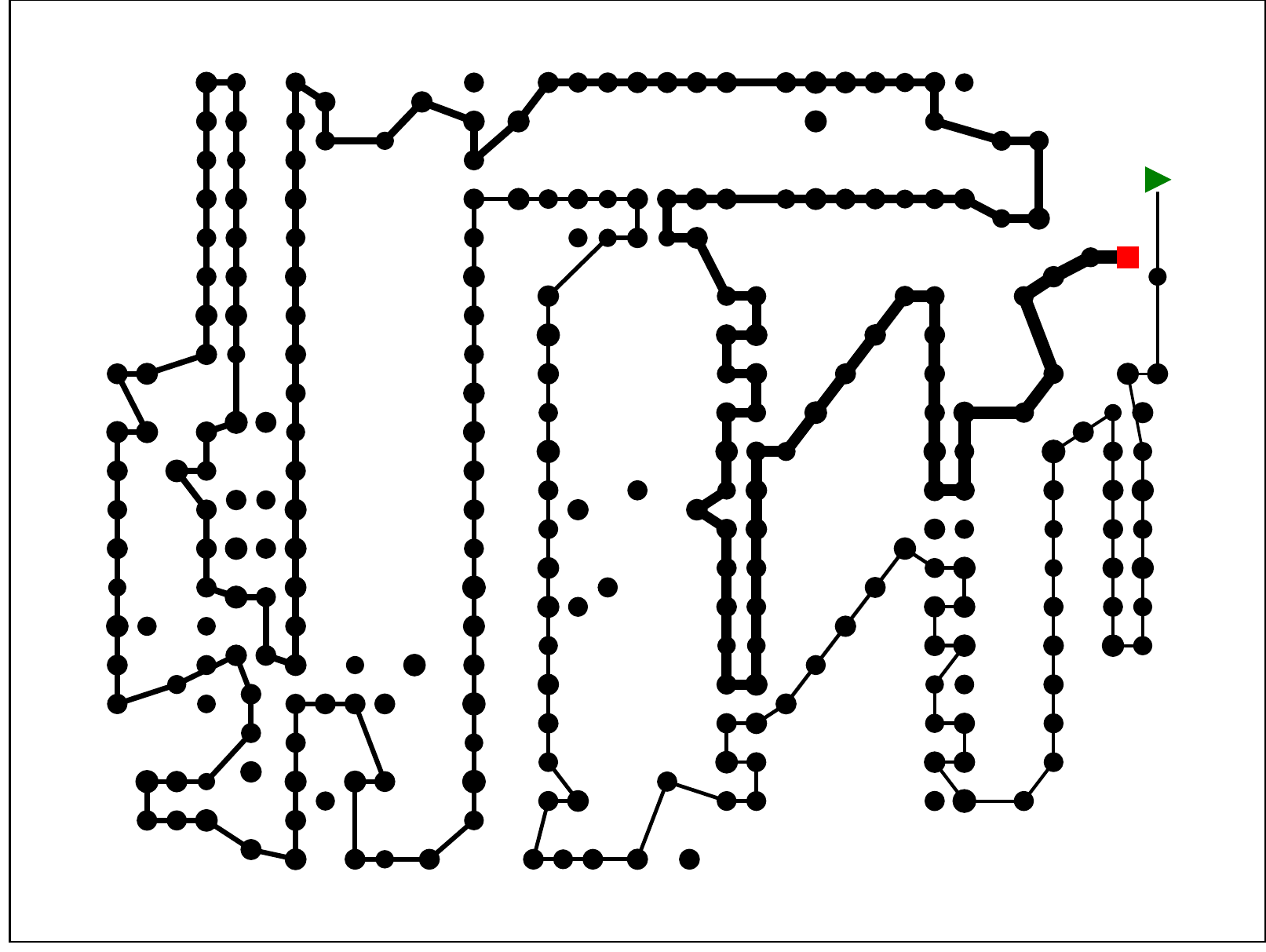}
 	}%
\caption{The graphical representation of the solutions found by ACO (left) and ACO\texttt{++} (right) for instances \textit{eil51\_10\_bsc\_01\_03} (top), \textit{pr107\_05\_usw\_10\_02} (middle), and \textit{a280\_10\_unc\_01\_03} (bottom). The initial and final cities are highlighted using a green triangle and a red square, respectively. The lines connecting pairs of cities represent the route traveled by the thief, with the increasing line thickness corresponding to the thief's knapsack weight.}
\label{fig:plot_solutions}
\end{figure*}

In Table~\ref{table:solution_info}, we summarize the results obtained with a closer analysis of the solutions found by ACO and ACO\texttt{++}. For each TSP base instance ({\tt XXX}) and number of items per city ({\tt YY}), which resulted in 27 instances each, we show averaged information concerning all the best solutions achieved by both approaches. Column $\mathcal{D}$ shows the ratio between the total distance traveled and the number of cities visited by the thief, while columns \textit{\%T} and \textit{\%W} report, respectively, the percentage spent of the time limit and the percentage used of the knapsack capacity. If values in these last two columns are close to 100\%, then these indicate limiting factors. Note that both algorithms have a similar use of the time limit. On the other hand, the solutions found by ACO\texttt{++} have used more the knapsack capacity, especially for instances with more cities and items. From the values in column $\mathcal{D}$, we can understand this behavior. Note that the ratio between the total distance traveled and the number of cities visited of the solutions found by ACO is higher than those found by ACO\texttt{++}. Note that the solutions found by ACO have a ratio between the total distance traveled and the number of cities visited higher than those solutions found by ACO\texttt{++}, which indicates that ACO has found the most spread-out routes and/or with more edge crossings. Therefore, as the routes found by ACO\texttt{++} are more condensed and/or efficient, the thief is able to travel more effectively and, consequently, uses better the knapsack capacity, thus managing to collect a better set of items. To illustrate this behavior, Figure~\ref{fig:plot_solutions} shows for some instances, where the resulting quality differs significantly, the best solution found by each algorithm. 

In summary, we can see that our ACO\texttt{++} has been able to find significantly more efficient routes, which allows achieving better packing plans, and, consequently, achieving higher profits.
 
\begin{table}[!ht]
\centering
\small
\caption{Information on the structure of the best solutions found.
$\mathcal{D}$ is the ratio between the total distance traveled and the number of cities visited; \textit{\%T} and \textit{\%W} denote the percentage spent of the time limit and the percentage used of the knapsack capacity.}
\setlength{\tabcolsep}{0pt}
\begin{tabular*}{\hsize}{@{}@{\extracolsep{\fill}}ccrrrlrrr@{}}
\toprule
\multirow{2}{*}{TSP base} & \multirow{2}{*}{Number of items} & \multicolumn{ 3}{c}{\multirow{2}{*}{ACO}} &  & \multicolumn{ 3}{c}{\multirow{2}{*}{ACO\texttt{++}}} \\ 
\multicolumn{1}{c}{\multirow{2}{*}{(\tt XXX)}}  & \multicolumn{1}{c}{\multirow{2}{*}{per city (\tt YY)}} & \multicolumn{ 3}{c}{} &  & \multicolumn{ 3}{c}{} \\
\cmidrule{3-5} \cmidrule{7-9} 
&  & \multicolumn{1}{c}{$\mathcal{D}$} & \multicolumn{1}{c}{\%T} & \multicolumn{1}{c}{\%W} & \multicolumn{1}{c}{} & \multicolumn{1}{c}{$\mathcal{D}$} & \multicolumn{1}{c}{\%T} & \multicolumn{1}{c}{\%W}\\
\midrule
eil51 & 01 & 10.91 & 98.60 & 78.06 &  & 10.46 & 99.00 & 79.55 \\ 
 & 03 & 9.46 & 99.60 & 83.85 &  & 8.63 & 99.55 & 85.34 \\ 
 & 05 & 9.26 & 99.62 & 83.39 &  & 8.53 & 99.78 & 85.48 \\ 
 & 10 & 9.12 & 99.84 & 85.26 &  & 8.20 & 99.88 & 86.57 \\[1mm]
pr107 & 01 & 718.92 & 99.66 & 79.58 &  & 680.85 & 99.74 & 81.82 \\ 
 & 03 & 498.71 & 99.85 & 81.78 &  & 476.67 & 99.85 & 83.84 \\ 
 & 05 & 471.77 & 99.93 & 83.22 &  & 445.23 & 99.95 & 83.79 \\ 
 & 10 & 449.09 & 99.95 & 84.95 &  & 417.47 & 99.93 & 84.40 \\[1mm] 
a280 & 01 & 16.52 & 99.61 & 79.57 &  & 14.23 & 99.80 & 83.94 \\ 
 & 03 & 12.60 & 99.74 & 81.83 &  & 10.45 & 99.76 & 85.68 \\ 
 & 05 & 11.84 & 99.95 & 82.72 &  & 9.61 & 99.93 & 86.27 \\ 
 & 10 & 11.17 & 99.92 & 83.02 &  & 9.22 & 99.92 & 86.26 \\[1mm] 
dsj1000 & 01 & 44632.08 & 74.72 & 79.31 &  & 37015.71 & 72.08 & 86.16 \\ 
 & 03 & 26635.61 & 99.90 & 82.91 &  & 18943.46 & 99.89 & 89.57 \\ 
 & 05 & 25648.23 & 99.85 & 82.43 &  & 18064.09 & 99.82 & 89.91 \\ 
 & 10 & 23795.22 & 99.92 & 84.03 &  & 17700.59 & 99.68 & 90.35 \\ 
\bottomrule
\end{tabular*}
\label{table:solution_info}
\end{table}

\subsection{Solving the classical Orienteering Problem}

One can note that the classical Orienteering Problem (OP)~\cite{GoLeVo87} is a subproblem of the ThOP, i.e., its definition is contained within the definition of the ThOP. Indeed, if ($i$) there is only one item per city $(|I_{i}| = 1,\;\forall i \in C \setminus \{1, n\})$, ($ii$)~the thief travels at a constant speed $(v_{max} = v_{min})$, and ($iii$) their knapsack has unlimited capacity $(W = \infty)$, we have the same constraints and objective on both problems. Therefore, we can use any ThOP algorithm for solving the classic OP, simply by \textit{fixing} some information in the ThOP instances in order to convert them into OP instances.

To investigate the efficiency of BRKGA, ILS, ACO, and ACO\texttt{++} algorithms proposed for the ThOP in solving the OP, we have adapted the ThOP instances that have only one item per city, setting their $v_{max} = v_{min} = 1$ and $W$ to a large number so that the knapsack capacity is greater than the sum of all item profits. These instances will be identified as {\tt XXX\_YY\_ZZZ\_WW\_TT}, as already mentioned in Section~\ref{section:benchmarking_instances}, with the emphasis now that {\tt WW} = \textit{inf}.

For our analysis, we have performed 30 independent runs per instance and algorithm. Then, for each instance, we have taken the average objective value obtained by each algorithm to compute the ratio between that value and the optimal objective value of that instance. We have determined the optimal value for each instance from the branch-and-cut algorithm proposed by Fischetti~et~al.~\cite{fischetti1998solving}, which has been executed without time limitation. In Table~\ref{table:op_instances}, we show all computed ratios, where the highest ratio for each instance is highlighted in bold. Note that the higher the ratio, the higher the average performance of that particular ThOP algorithm in solving the OP instance.

\begin{table}
\centering
\small
\caption{Ratios between the average objective value obtained by each algorithm and the optimal objective value for OP instances.}
\setlength{\tabcolsep}{0pt}
\begin{tabular*}{\hsize}{@{}@{\extracolsep{\fill}}lccccc@{}}
\toprule
\multicolumn{1}{c}{Instance} & & \multicolumn{1}{c}{ILS} & \multicolumn{1}{c}{BRKGA} & \multicolumn{1}{c}{ACO} & \multicolumn{1}{c}{ACO\texttt{++}} \\
\midrule
\texttt{eil51\_01\_bsc\_inf\_01} & & 0.922 & 0.953 & 0.921 & \textBF{0.965} \\
\texttt{eil51\_01\_bsc\_inf\_02} & & 0.912 & 0.961 & 0.977 & \textBF{0.995} \\
\texttt{eil51\_01\_bsc\_inf\_03} & & 0.879 & 0.964 & 0.968 & \textBF{0.988} \\
\texttt{eil51\_01\_unc\_inf\_01} & & 0.872 & 0.963 & 0.965 & \textBF{0.997} \\
\texttt{eil51\_01\_unc\_inf\_02} & & 0.856 & 0.956 & 0.989 & \textBF{0.996} \\
\texttt{eil51\_01\_unc\_inf\_03} & & 0.905 & 0.979 & 0.999 & \textBF{1.000} \\
\texttt{eil51\_01\_usw\_inf\_01} & & \textBF{1.000} & 0.989 & \textBF{1.000} & \textBF{1.000} \\
\texttt{eil51\_01\_usw\_inf\_02} & & 0.979 & 0.996 & 0.969 & \textBF{1.000} \\
\texttt{eil51\_01\_usw\_inf\_03} & & 0.917 & 0.958 & 0.966 & \textBF{1.000} \\[1mm]
\texttt{pr107\_01\_bsc\_inf\_01} & & 0.721 & 0.852 & 0.923 & \textBF{0.940} \\
\texttt{pr107\_01\_bsc\_inf\_02} & & 0.709 & 0.892 & 0.967 & \textBF{0.999} \\
\texttt{pr107\_01\_bsc\_inf\_03} & & 0.791 & 0.969 & \textBF{1.000} & \textBF{1.000} \\
\texttt{pr107\_01\_unc\_inf\_01} & & 0.689 & 0.954 & 0.935 & \textBF{0.955} \\
\texttt{pr107\_01\_unc\_inf\_02} & & 0.668 & 0.904 & 0.924 & \textBF{0.961} \\
\texttt{pr107\_01\_unc\_inf\_03} & & 0.718 & 0.915 & 0.999 & \textBF{1.000} \\
\texttt{pr107\_01\_usw\_inf\_01} & & 0.861 & \textBF{0.985} & 0.945 & 0.965 \\
\texttt{pr107\_01\_usw\_inf\_02} & & 0.732 & 0.943 & 0.961 & \textBF{0.979} \\
\texttt{pr107\_01\_usw\_inf\_03} & & 0.708 & 0.908 & 0.940 & \textBF{0.971} \\[1mm]
\texttt{a280\_01\_bsc\_inf\_01} & & 0.537 & 0.683 & 0.781 & \textBF{0.882} \\
\texttt{a280\_01\_bsc\_inf\_02} & & 0.516 & 0.655 & 0.869 & \textBF{0.954} \\
\texttt{a280\_01\_bsc\_inf\_03} & & 0.533 & 0.675 & 0.981 & \textBF{0.997} \\
\texttt{a280\_01\_unc\_inf\_01} & & 0.484 & 0.639 & 0.774 & \textBF{0.927} \\
\texttt{a280\_01\_unc\_inf\_02} & & 0.476 & 0.642 & 0.932 & \textBF{0.973} \\
\texttt{a280\_01\_unc\_inf\_03} & & 0.522 & 0.709 & \textBF{1.000} & \textBF{1.000} \\
\texttt{a280\_01\_usw\_inf\_01} & & 0.637 & 0.750 & 0.833 & \textBF{0.956} \\
\texttt{a280\_01\_usw\_inf\_02} & & 0.500 & 0.657 & 0.814 & \textBF{0.932} \\
\texttt{a280\_01\_usw\_inf\_03} & & 0.476 & 0.627 & 0.828 & \textBF{0.949} \\[1mm]
\texttt{dsj1000\_01\_bsc\_inf\_01} & & 0.669 & 0.882 & \textBF{1.000} & \textBF{1.000} \\
\texttt{dsj1000\_01\_bsc\_inf\_02} & & 0.836 & 0.952 & \textBF{1.000} & \textBF{1.000} \\
\texttt{dsj1000\_01\_bsc\_inf\_03} & & 0.965 & 0.995 & \textBF{1.000} & \textBF{1.000} \\
\texttt{dsj1000\_01\_unc\_inf\_01} & & 0.589 & 0.904 & \textBF{1.000} & \textBF{1.000} \\
\texttt{dsj1000\_01\_unc\_inf\_02} & & 0.759 & 0.961 & \textBF{1.000} & \textBF{1.000} \\
\texttt{dsj1000\_01\_unc\_inf\_03} & & 0.956 & 0.995 & \textBF{1.000} & \textBF{1.000} \\
\texttt{dsj1000\_01\_usw\_inf\_01} & & 0.272 & 0.656 & \textBF{1.000} & \textBF{1.000} \\
\texttt{dsj1000\_01\_usw\_inf\_02} & & 0.363 & 0.770 & \textBF{1.000} & \textBF{1.000} \\
\texttt{dsj1000\_01\_usw\_inf\_03} & & 0.438 & 0.882 & \textBF{1.000} & \textBF{1.000} \\
\midrule
Average & & 0.705 & 0.863 & 0.949 & \textBF{0.980} \\ 
\bottomrule
\end{tabular*}
\label{table:op_instances}
\end{table}

From Table~\ref{table:op_instances}, we can verify the same behavior among the algorithms when solving OP instances as seen in the previous experiments: better efficiency of the algorithms based on the ACO strategy, with a significant better performance in our ACO\texttt{++} algorithm. Note that our algorithm is able to find high-quality solutions for all instances. For almost all instances, the ratio between the objective values of solutions found by the ACO\texttt{++} and their optimal values is larger than 0.95. On average, our algorithm has reached a ratio of 0.98. Note that for all instances with the TSP base {\it dsj1000}, both algorithms based on ACO has found the optimal solutions (ratio = 1.00). Analyzing such solutions, we see that the thief is able to visit all cities and, consequently, steal all items. This has occurred because without the speed reduction and with an unlimited knapsack, these instances become easier once the thief has enough time and space in their knapsack to steal and carry all items along a path that does not need to be very optimized. Anyway, it is noteworthy that the other algorithms have not been able to find such solutions.

\clearpage
\section{Conclusions}
\label{sec:conclusions}

In this article, we have proposed an improvement to a swarm-based approach to the academic Thief Orienteering Problem (ThOP): we have combined a heuristic approach based on Ant Colony Optimization with a randomized packing heuristic and with local searches. Using extensive tuning on groups of instances, we have studied the effects of our algorithmic components. Furthermore, we have evaluated the performance of the algorithm on the complete set of instances available in the literature. The experiments show that our solution strategy is able to find better solutions with large improvements when compared to the other solution methods proposed for the problem. Based on our analysis, the efficiency of our algorithm is due to the fact that ants have been able to find more efficient routes, which has allowed our packing heuristic to select a better set of items. In addition, we have shown that our algorithm is able to find high-quality solutions for the classic Orienteering Problem without any modification in its design.

For future research, we may investigate a variant of the current problem that generalizes to multiple thieves. Another interesting direction will be to approach the problem in a bi-objective version, where are to maximize the total profit and minimize the total distance traveled. By combining both foregoing directions, it will create a very interesting, challenging, and general problem with potential applications in real-world scenarios with routing problems under time-dependent limitations.

\begin{acknowledgements}
The authors would like to thank the following institutions and funding bodies: 
Coordena\c{c}\~{a}o de A\-per\-fei\-\c{c}o\-a\-men\-to de Pessoal de N\'{i}vel Superior - Brazil (CAPES) - Finance code 001; Funda\c{c}\~{a}o de Amparo \`{a} Pesquisa do Estado de Minas Gerais (FAPEMIG); Conselho Nacional de Desenvolvimento Cient\'{i}fico e Tecnol\'{o}gico (CNPq); Universidade Federal de Ouro Preto (UFOP); Universidade Federal de Vi\c{c}osa (UFV); and Australian Research Council Project DP200102364.
\end{acknowledgements}

%
%

\FloatBarrier
%
%
\bibliographystyle{spmpsci}      

\bibliography{bibfile}   

\begin{thebibliography}{10}
\providecommand{\url}[1]{{#1}}
\providecommand{\urlprefix}{URL }
\expandafter\ifx\csname urlstyle\endcsname\relax
  \providecommand{\doi}[1]{DOI~\discretionary{}{}{}#1}\else
  \providecommand{\doi}{DOI~\discretionary{}{}{}\begingroup
  \urlstyle{rm}\Url}\fi

\bibitem{aarts2003local}
Aarts, E., Aarts, E.H., Lenstra, J.K.: Local search in combinatorial
  optimization.
\newblock Princeton University Press (2003)

\bibitem{birattari2010f}
Birattari, M., Yuan, Z., Balaprakash, P., St{\"u}tzle, T.: F-race and iterated
  f-race: An overview.
\newblock In: Experimental methods for the analysis of optimization algorithms,
  pp. 311--336. Springer (2010)

\bibitem{bonyadi2013travelling}
Bonyadi, M.R., Michalewicz, Z., Barone, L.: The travelling thief problem: The
  first step in the transition from theoretical problems to realistic problems.
\newblock In: IEEE Congress on Evolutionary Computation, pp. 1037--1044. IEEE
  (2013)

\bibitem{bonyadi2019evolutionary}
Bonyadi, M.R., Michalewicz, Z., Wagner, M., Neumann, F.: Evolutionary
  computation for multicomponent problems: opportunities and future directions.
\newblock In: Optimization in Industry, pp. 13--30. Springer (2019)

\bibitem{chagas2020ants}
Chagas, J.B., Wagner, M.: Ants can orienteer a thief in their robbery.
\newblock Operations Research Letters \textbf{48}(6), 708 -- 714 (2020)

\bibitem{chand2016fast}
Chand, S., Wagner, M.: Fast heuristics for the multiple traveling thieves
  problem.
\newblock In: Genetic and Evolutionary Computation Conference (GECCO), pp.
  293--300. ACM (2016)

\bibitem{conf/hcomp/ChenCGMDC14}
Chen, C., Cheng, S.F., Gunawan, A., Misra, A., Dasgupta, K., Chander, D.:
  Traccs: A framework for trajectory-aware coordinated urban crowd-sourcing.
\newblock In: J.P. Bigham, D.C. Parkes (eds.) Second AAAI Conference on Human
  Computation and Crowdsourcing (HCOMP). AAAI (2014).
\newblock \urlprefix\url{http://www.aaai.org/Library/HCOMP/hcomp14contents.php}

\bibitem{dorigo1999ant}
Dorigo, M., Di~Caro, G.: Ant colony optimization: a new meta-heuristic.
\newblock In: IEEE Congress on Evolutionary Computation (CEC), vol.~2, pp.
  1470--1477. IEEE (1999)

\bibitem{faeda2020genetic}
Fa{\^e}da, L.M., Santos, A.G.: A genetic algorithm for the thief orienteering
  problem.
\newblock In: 2020 IEEE Congress on Evolutionary Computation (CEC), pp. 1--8.
  IEEE (2020)

\bibitem{faulkner2015approximate}
Faulkner, H., Polyakovskiy, S., Schultz, T., Wagner, M.: Approximate approaches
  to the traveling thief problem.
\newblock In: Genetic and Evolutionary Computation Conference (GECCO), pp.
  385--392. ACM (2015)

\bibitem{fischetti1998solving}
Fischetti, M., Gonzalez, J.J.S., Toth, P.: Solving the orienteering problem
  through branch-and-cut.
\newblock INFORMS Journal on Computing \textbf{10}(2), 133--148 (1998)

\bibitem{GamrathEtal2020ZR}
Gamrath, G., Anderson, D., Bestuzheva, K., Chen, W.K., Eifler, L., Gasse, M.,
  Gemander, P., Gleixner, A., Gottwald, L., Halbig, K., Hendel, G., Hojny, C.,
  Koch, T., Le~Bodic, P., Maher, S.J., Matter, F., Miltenberger, M.,
  M{\"u}hmer, E., M{\"u}ller, B., Pfetsch, M.E., Schl{\"o}sser, F., Serrano,
  F., Shinano, Y., Tawfik, C., Vigerske, S., Wegscheider, F., Weninger, D.,
  Witzig, J.: {The SCIP Optimization Suite 7.0}.
\newblock ZIB-Report 20-10, Zuse Institute Berlin (2020).
\newblock \urlprefix\url{http://nbn-resolving.de/urn:nbn:de:0297-zib-78023}

\bibitem{GoLeVo87}
Golden, B.L., Levy, L., Vohra, R.: The orienteering problem.
\newblock Naval Research Logistics \textbf{34}, 307--318 (1987)

\bibitem{GUNAWAN2016315}
Gunawan, A., Lau, H.C., Vansteenwegen, P.: Orienteering problem: A survey of
  recent variants, solution approaches and applications.
\newblock European Journal of Operational Research \textbf{255}(2), 315 -- 332
  (2016)

\bibitem{iori2010routing}
Iori, M., Martello, S.: Routing problems with loading constraints.
\newblock Top \textbf{18}(1), 4--27 (2010)

\bibitem{KIM2020106808}
Kim, H., Kim, B.I., jin Noh, D.: The multi-profit orienteering problem.
\newblock Computers and Industrial Engineering \textbf{149}, 106808 (2020)

\bibitem{lopez2016irace}
L{\'o}pez-Ib{\'a}{\~n}ez, M., Dubois-Lacoste, J., C{\'a}ceres, L.P., Birattari,
  M., St{\"u}tzle, T.: The irace package: Iterated racing for automatic
  algorithm configuration.
\newblock Operations Research Perspectives \textbf{3}, 43--58 (2016)

\bibitem{PySCIPOpt2016}
Maher, S., Miltenberger, M., Pedroso, J.P., Rehfeldt, D., Schwarz, R., Serrano,
  F.: {PySCIPOpt}: Mathematical programming in python with the {SCIP}
  optimization suite.
\newblock In: Mathematical Software {\textendash} {ICMS} 2016, pp. 301--307.
  Springer International Publishing (2016).
\newblock \doi{10.1007/978-3-319-42432-3_37}

\bibitem{Neumann2019fptasPWT}
Neumann, F., Polyakovskiy, S., Skutella, M., Stougie, L., Wu, J.: A fully
  polynomial time approximation scheme for packing while traveling.
\newblock In: Y.~Disser, V.S. Verykios (eds.) Algorithmic Aspects of Cloud
  Computing, pp. 59--72. Springer (2019)

\bibitem{journals/transci/OrlisBRD20}
Orlis, C., Bianchessi, N., Roberti, R., Dullaert, W.: The team orienteering
  problem with overlaps: An application in cash logistics.
\newblock Transportation Science \textbf{54}(2), 470--487 (2020)

\bibitem{polyakovskiy2014comprehensive}
Polyakovskiy, S., Bonyadi, M.R., Wagner, M., Michalewicz, Z., Neumann, F.: A
  comprehensive benchmark set and heuristics for the traveling thief problem.
\newblock In: Genetic and Evolutionary Computation Conference (GECCO), pp.
  477--484. ACM (2014)

\bibitem{polyakovskiy2015packing}
Polyakovskiy, S., Neumann, F.: Packing while traveling: Mixed integer
  programming for a class of nonlinear knapsack problems.
\newblock In: International Conference on AI and OR Techniques in Constriant
  Programming for Combinatorial Optimization Problems (CPAIOR), pp. 332--346.
  Springer (2015)

\bibitem{santos2018thief}
Santos, A.G., Chagas, J.B.: The thief orienteering problem: Formulation and
  heuristic approaches.
\newblock In: IEEE Congress on Evolutionary Computation (CEC), pp. 1191--1199.
  IEEE (2018)

\bibitem{stutzle2000max}
St{\"u}tzle, T., Hoos, H.H.: Max--min ant system.
\newblock Future generation computer systems \textbf{16}(8), 889--914 (2000)

\bibitem{toth1990knapsack}
Toth, P., Martello, S.: Knapsack problems: Algorithms and computer
  implementations.
\newblock Wiley (1990)

\bibitem{TRACHANATZI2020100712}
Trachanatzi, D., Rigakis, M., Marinaki, M., Marinakis, Y.: A firefly algorithm
  for the environmental prize-collecting vehicle routing problem.
\newblock Swarm and Evolutionary Computation p. 100712 (2020)

\bibitem{wagner2016stealing}
Wagner, M.: Stealing items more efficiently with ants: a swarm intelligence
  approach to the travelling thief problem.
\newblock In: International Conference on Swarm Intelligence (ANTS), pp.
  273--281. Springer (2016)

\bibitem{wagner2018case}
Wagner, M., Lindauer, M., M{\i}s{\i}r, M., Nallaperuma, S., Hutter, F.: A case
  study of algorithm selection for the traveling thief problem.
\newblock Journal of Heuristics \textbf{24}(3), 295--320 (2018)

\bibitem{wu2017exact}
Wu, J., Wagner, M., Polyakovskiy, S., Neumann, F.: Exact approaches for the
  travelling thief problem.
\newblock In: Asia-Pacific Conference on Simulated Evolution and Learning, pp.
  110--121. Springer (2017)

\end{thebibliography}

\end{document}